\documentclass[twoside,11pt]{article}

\usepackage{jmlr2e}
\usepackage[font=small]{caption}
\usepackage{amsmath}
\usepackage{amssymb}

\usepackage{subcaption} 
\usepackage{booktabs}
\usepackage[ruled,vlined]{algorithm2e}
\usepackage{algpseudocode}
\usepackage{scalerel}
\usepackage{verbatim}
\usepackage{commath}
\SetCommentSty{mycommfont}

\algdef{SE}{Begin}{End}{\textbf{begin}}{\textbf{end}}

\DeclareMathOperator{\tensorize}{ten}
\DeclareMathAlphabet\mathcalbf{OMS}{cmsy}{b}{n}

\ShortHeadings{Tensor Networks for Multi-Modal Non-Euclidean Data}{Xu, Konstantinidis \& Mandic}
\firstpageno{1}

\begin{document}


\title{Tensor Networks for Multi-Modal Non-Euclidean Data}

\author{\name Yao Lei Xu \email yao.xu15@imperial.ac.uk \AND
       \name Kriton Konstantinidis \email k.konstantinidis19@imperial.ac.uk \AND
       \name Danilo P. Mandic \email d.mandic@imperial.ac.uk\\
       \addr Department of Electrical and Electronic Engineering\\
       Imperial College London\\
       London SW7 2AZ, U.K.}

\editor{}

\maketitle

\begin{abstract}
Modern data sources are typically of large scale and multi-modal natures, and acquired on irregular domains, which poses serious challenges to traditional deep learning models. These issues are partially mitigated by either extending existing deep learning algorithms to irregular domains through graphs, or by employing tensor methods to alleviate the computational bottlenecks imposed by the Curse of Dimensionality. To simultaneously resolve both these issues, we introduce a novel \textit{Multi-Graph Tensor Network} (MGTN) framework, which leverages on the desirable properties of graphs, tensors and neural networks in a physically meaningful and compact manner. This equips MGTNs with the ability to exploit local information in irregular data sources at a drastically reduced parameter complexity, and over a range of learning paradigms such as regression, classification and reinforcement learning. The benefits of the MGTN framework, especially its ability to avoid overfitting through the inherent low-rank regularization properties of tensor networks, are demonstrated through its superior performance against competing models in the individual tensor, graph, and neural network domains.
\end{abstract}


\section{Introduction}

Deep learning has been at the core of machine learning research for over a decade and has proved successful in a number of areas, notably computer vision and natural language processing ~\citep{zhang2018survey}. However, as we enter the era of Big Data, the associated multi-modal and irregular nature of data is posing serious challenges to traditional learning systems; these include the sheer volume, variety, veracity and velocity of modern data sources ~\citep{Cichocki2014}. It is therefore both necessary and timely to create new deep learning frameworks, that are particularly suitable for handling such irregular and multi-modal data.

Some of the most successful approaches to data analytics on irregular domains belong to graph data analysis techniques, because of their ability to provide insights into both the data acquisition process and their generative mechanisms ~\citep{shuman2013emerging}. Indeed, by virtue of their ability to account for the underlying data structure, graph-based learning algorithms have proved advantageous in applications where the graph is known a-priori ~\citep{wu2020comprehensive}. On the other hand, when it comes to exceedingly large multi-modal data, tensor-based methods have demonstrated their potential in areas including multi-modal learning ~\citep{7038247}, compression of large-dimensional data ~\citep{cichocki2016tensor}, and interpretability of neural networks ~\citep{cohen2016expressive}. In particular, tensor decompositions (TD) and tensor networks (TN) leverage on the multi-modality inherent to many Big Data applications to compress large-dimensional data while preserving their structure and interpretability. In this way, tensors help bypass the bottlenecks imposed by the Curse of Dimensionality ~\citep{Cichocki2014}.

To address the issues associated with the interpretability and irregular data domains in deep learning scenarios, we introduce \textit{Multi-Graph Tensor Networks} (MGTN) ~\citep{xu2020multigraph}, a general framework that fully exploits the advantages of both graphs and tensors in a deep learning setting. The proposed framework is shown to be capable of handling irregular data residing on multiple graph domains, while simultaneously leveraging the compression and regularization properties of tensor networks, thus simultaneously enhancing modelling power and reducing complexity.

The proposed framework is validated over four different tasks spanning diverse data natures and degrees of complexity, and across the regression, reinforcement learning and classification learning paradigms. First, we show its applicability on the task of Foreign Exchange (FOREX) algorithmic trading, a notoriously challenging paradigm characterized by highly irregular and noisy data ~\citep{de2020machine}. Next, the MGTN is employed for the task of mental state classification based on Electroencephalogram (EEG) recordings. We then demonstrate the advantages of MGTN in temperature forecasting across different cities in United States, and finally investigate the performance of the proposed framework in predicting air quality.  
By combining the advantages of graphs, tensors and neural networks, the proposed MGTN framework is shown to yield highly superior performance against the competing models in their respective graph, tensor and neural network domains, across all experiments considered.
The superior performance of the proposed framework in these four paradigms suggests the potential of MGTN in a range of other application domains, including social networks, communication networks, and cognitive neuroscience, to name but a few.

The rest of the paper is organized as follows. In Section \ref{sec:related_works}, we first discuss related work and elaborate on the differences and advantages of the proposed framework, before presenting the theoretical background necessary to follow this work in Section \ref{sec:Prelim}. Next, Section \ref{sec:GTN} introduces the MGTN framework, followed by an in-depth analysis of four experimental setups and extensive comparisons between the performance of different competing models and the proposed framework in Section \ref{sec:exp}. Finally, we conclude with promising future research directions and potential MGTN application domains.
 
\section{Related Work} \label{sec:related_works}

Since the introduction of deep learning on graph domains ~\citep{7038247}, a number of different Graph Neural Network (GNN) models have been introduced, including Graph Convolutional Neural Networks (GCNs) ~\citep{kipf2016semi,NIPS2016_04df4d43}, as well as GNNs for sequential data ~\citep{DBLP:journals/corr/LiTBZ15}, to name but a few. For more details on different GNN models we refer the reader to ~\citep{wu2020comprehensive}.
The major difference between the proposed framework and the existing approaches is that the MGTN model is designed to cater for the multi-modality of data on non-Euclidean domains, while being fully described in a structure-aware Tensor Network format.

A more closely related approach to our work is ~\citep{NIPS2017_2eace51d}, where the authors develop a Multi-Graph Neural Network, employed for matrix completion for a recommender system paradigm. However, the proposed models are limited to application domains where the dimensionality of the samples is exactly the same as the number of nodes in the considered graphs, which is often too restrictive for Big Data applications. To deal with this issue, due to the inherent multi-modal structure of tensor networks, the MGTN architecture has the ability to generalize the multi-graph filtering operation to data structures of any dimensionality.

Despite tremendous progress in deep learning, graph, and tensor research, the full potential arising from the combination of these individual fields has only begun to be explored, with very few existing works along these lines. One such recent approach is the \textit{Recurrent Graph Tensor Network} (RGTN) ~\citep{xu2020recurrent}, which provides a framework for modelling multi-modal sequential data through a unifying account of the expressive power of graphs and tensor networks. The RGTN model has been introduced for sequential data and is therefore only defined on a single graph domain, which is often impractical for Big Data applications. On the other hand, the proposed MGTN can operate on any number of graphs and across the learning paradigms, including regression and classification.

\section{Preliminaries}
\label{sec:Prelim}

\subsection{Tensors and Tensor Networks}

A real-valued tensor is a multidimensional array, denoted by a calligraphic font, e.g., $\mathcalbf{X} \in\mathbb{R}^{I_1\times\dots\times I_N}$, where $N$ is the order of the tensor and $I_n$ ($1 \leq n \leq N$) the size of its $n$\textsuperscript{th} mode. Matrices (denoted by bold capital letters, e.g., $\mathbf{X}\in\mathbb{R}^{I_1\times I_2}$) can be seen as order-2 tensors ($N=2$), vectors (denoted by bold lower-case letters, e.g., $\mathbf{x}\in\mathbb{R}^{I}$) can be seen as order-1 tensors ($N=1$), and scalars (denoted by lower-case letters, e.g., $x\in\mathbb{R}$) are tensors of order $N=0$. A specific entry of a tensor $\mathcalbf{X}\in\mathbb{R}^{I_1\times\dots\times I_N}$ is given by $x_{i_1,\dots,i_N}\in\mathbb{R}$. The tensor indices in this paper are grouped according to the Little-Endian convention ~\citep{Dolgov2014}.

\textbf{Kronecker Product} A (left) Kronecker product between two tensors, $\mathcalbf{A} \in \mathbb{R}^{I_1 \times \cdots \times I_N}$ and $\mathcalbf{B} \in \mathbb{R}^{J_1 \times \cdots \times J_N}$, denoted by $\otimes$, yields a tensor  $\mathcalbf{C} \in \mathbb{R}^{I_1 J_1 \times \cdots \times I_N J_N}$, of the same order, with entries $c_{\overline{i_1j_1},\ldots,\overline{i_Nj_N}} = a_{i_1, \ldots, i_N} b_{j_1, \ldots, j_N}$, where $\overline{i_n j_n} = j_n + (i_n - 1) J_n$ ~\citep{Cichocki2014}.

\textbf{Matricization and Tensorization} The mode-$n$ \textit{matricization} of a tensor $\mathcalbf{X}\in\mathbb{R}^{I_1\times\cdots\times I_N}$ reshapes the multidimensional array into a matrix $\mathbf{X}_{(n)}\in\mathbb{R}^{I_n\times I_1I_2\cdots I_{n-1}I_{n+1}\cdots I_N}$ with entry, $(x_{(n)})_{i_n,\overline{i_1\dots i_{n-1}i_{n+1}\dots i_N}}=x_{i_1,\dots,i_N}$. The inverse process, \textit{tensorization}, is denoted by $\tensorize(\cdot)$.

\textbf{Tensor Contraction} An $(m,n)$-contraction ~\citep{Cichocki2014} denoted by $\times^m_n$, between an order-$N$ tensor $\mathcalbf{A} \in \mathbb{R}^{I_1\times \cdots \times I_n \times \cdots \times I_N}$ and an order-$M$ tensor $\mathcalbf{B}\in \mathbb{R}^{J_1\times \dots \times J_m \times \dots \times J_M} $, where $I_n = J_m$, yields, $\mathcalbf{C}\in \mathbb{R}^{I_1 \times \cdots \times I_{n-1} \times I_{n+1}  \times \cdots \times I_N \times J_1 \times \cdots \times J_{m-1} \times J_{m+1}  \times \cdots \times J_M}$, a third order-$(N+M-2)$ tensor, where each entry, $c_{i_1,\dots,i_{n-1}, i_{n+1}, \dots, i_N, j_1, \dots, j_{m-1}, j_{m+1}, \dots, j_M}$, is defined as, $c_{i_1,\dots,i_{n-1}, i_{n+1}, \dots, i_N, j_1, \dots, j_{m-1}, j_{m+1}, \dots, j_M} = \sum_{i_n=1}^{I_n} a_{i_1, \dots, i_{n-1}, i_n, i_{n+1}, \dots, i_N } b_{j_1, \dots, j_{m-1}, i_n, j_{m+1}, \dots, j_M}$. 

\begin{figure}[t]
	\centering
	\includegraphics[width=0.8\linewidth]{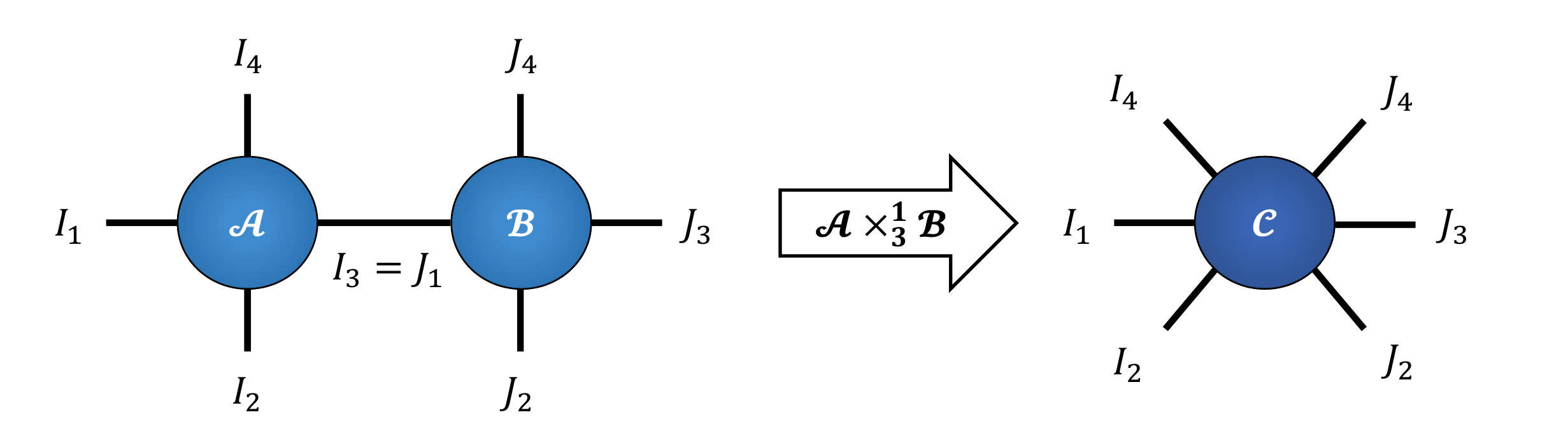}
	\vspace{-4mm}
	\caption{\textit{Tensor network representation of a contraction $\mathcalbf{A} \times_3^1 \mathcalbf{B}$ between tensors $\mathcalbf{A} \in \mathbb{R}^{I_1 \times I_2 \times I_3 \times I_4}$ and $\mathcalbf{B} \in \mathbb{R}^{J_1 \times J_2 \times J_3 \times J_4}$ over the modes with equal dimensions $I_3=J_1$.}}
	\label{fig:ContractionTN}
\end{figure}

\textbf{Tensor Networks}
A Tensor Network (TN) is a tensor architecture comprised of smaller-order core tensors which are connected by tensor contractions, whereby each tensor is represented as a node, while the number of edges that extends from that node corresponds to tensor order ~\citep{cichocki2016tensor}. If two nodes are connected through an edge, it represents a linear contraction between two tensors over modes of equal dimensions. Figure \ref{fig:ContractionTN} illustrates a tensor contraction operation using tensor network notation.

\begin{figure}[t]
	\centering
	\includegraphics[width=0.8\linewidth]{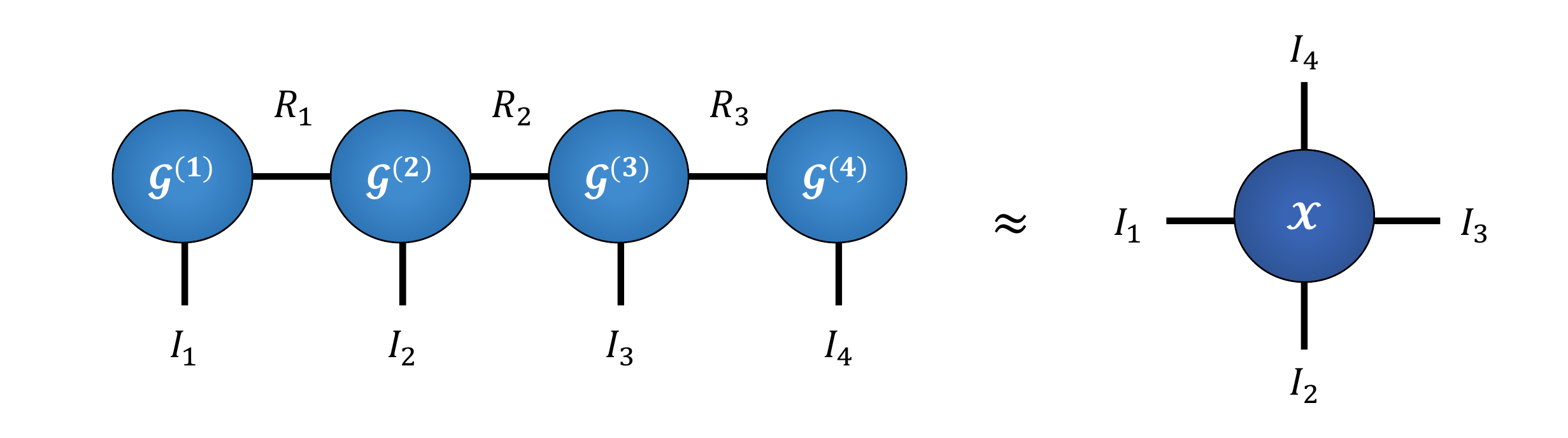}
	\vspace{-4mm}
	\caption{\textit{Tensor network representation of TT decomposition (TTD) for an order 4 tensor $\mathcalbf{X} \in \mathbb{R}^{I_1 \times I_2 \times I_3 \times I_4}$.}}
	\label{fig:TTDTN}
\end{figure}

\textbf{Tensor Decompositions} 
Special instances of tensor networks include those based on Tensor Decomposition (TD) methods, which approximate high-order, large-dimension tensors via contractions of smaller core tensors, therefore drastically reducing the computational complexity in tensor manipulation while preserving the data structure ~\citep{cichocki2016tensor}. We here consider the Tensor-Train decomposition (TTD) ~\citep{oseledets2011tensor}, a highly efficient TD method that can decompose a large order-$N$ tensor, $\mathcalbf{X} \in \mathbb{R}^{I_1 \times I_2 \times \cdots \times I_N}$, into interconnected smaller core tensors, $\mathcalbf{G}^{(n)} \in \mathbb{R}^{ R_{n-1} \times  I_n \times R_n }$, as $\mathcalbf{X} = \mathcalbf{G}^{(1)} \times^1_2 \mathcalbf{G}^{(2)} \times^1_3 \mathcalbf{G}^{(3)} \times^1_3 \cdots \times^1_3 \mathcalbf{G}^{(N)}$, where the set of $R_n$ for $n=0,\ldots,N$ and $R_0 = R_N = 1$ is referred to as the \textit{TT-rank}. The compression properties of TTD can be applied to significantly compress neural networks while maintaining comparable performance ~\citep{Novikov2015tnn}. An example of a TTD is shown in Figure \ref{fig:TTDTN}.


\subsection{Graph Signal Processing}
\label{sec:GSP}

A graph, $\mathcal{G} = \{\mathcal{V}, \mathcal{E}\}$, is defined by a set of $N$ vertices (or nodes) $\textit{v}_n \subset \mathcal{V}$ for $n = 1, \ldots , N$, and a set of edges connecting the $n^{th}$ and $m^{th}$ vertex $\textit{e}_{n,m} = (\textit{v}_n, \textit{v}_m) \in \mathcal{E}$, for $n=1,\ldots,N$ and $m=1,\ldots,N$. A signal on a given graph is a defined by a vector $\textbf{f} \in \mathbb{R} ^ {N}$ such that $\textbf{f}: \mathcal{V} \rightarrow \mathbb{R}$, which associates a signal value to every node on the graph ~\citep{stankovic2019graph}. A graph can be fully described in terms of its weighted adjacency matrix, $\textbf{A} \in \mathbb{R} ^ {N \times N}$, such that $\textit{a}_{n, m} > 0$ if $\textit{e}_{n,m} \in \mathcal{E}$, and $\textit{a}_{n, m} = 0$ if $\textit{e}_{n,m} \notin \mathcal{E}$. The adjacency matrix can also be represented in its normalized form as $\tilde{\textbf{A}} = \textbf{D}^{\frac{1}{2}} \textbf{A} \textbf{D}^{\frac{1}{2}}$, where  $\textbf{D} \in \mathbb{R} ^ {N \times N}$ is the diagonal degree matrix such that $d_{n,n} = \sum_m \textit{a}_{n,m}$.

The weighted adjacency matrix can be used as a shift operator to filter signals on graphs. Such a graph filter represents a linear combination of vertex-shifted graph signals, which captures graph information at a local level ~\citep{stankovic2019graphII}. For example, the operation $\textbf{g} = (\textbf{I} + \textbf{A})\textbf{f}$ produces a filtered signal, $\textbf{g} \in \mathbb{R}^{N}$, such that $g_n = f_n + \sum_{m \in \Omega_n} a_{n, m} f_m$, where $\Omega_n$ denotes the $1$-hop neighbours that are directly connected to the $n$-th node. For $M$ graph signals stacked in a matrix form as $\textbf{F} \in \mathbb{R}^{N \times M}$, the resulting graph filter can be compactly written as $\textbf{G} = (\textbf{I} + \textbf{A})\textbf{F}$ ~\citep{stankovic2019graphII}. 

\subsection{Recurrent Graph Tensor Networks}
\label{sec:RGTN}

A Recurrent Graph Tensor Network (RGTN) ~\citep{xu2020recurrent} models sequential data through a time-based, multi-linear graph filter in a tensor network format.

\textbf{The gRGTN Model}
A general RGTN (gRGTN) extracts a feature map $\textbf{Y} \in \mathbb{R}^{J_1 \times I_1}$ from sequential data, $\textbf{X} \in \mathbb{R}^{J_0 \times I_1}$, via the forward pass given by  $\textbf{Y} = \sigma(\mathcalbf{R} \times_{3,4}^{1,2} \textbf{W}^{(x)} \times_2^1 \textbf{X})$, where $\textbf{W}^{(x)} \in \mathbb{R}^{J_1 \times J_0}$ is the input weight matrix and $\mathcalbf{R} \in \mathbb{R}^{J_1 \times I_1 \times J_1 \times I_1}$ is the multi-linear time-graph filter. Specifically, $\mathcalbf{R}$ is defined as $\mathcalbf{R} = \tensorize(\textbf{I} + (\textbf{A} \otimes \textbf{W}^{(r)}))$, where $\textbf{I} \in \mathbb{R}^{J_1 I_1 \times J_1 I_1}$ is the identity matrix, $\textbf{A} \in \mathbb{R}^{I_1 \times I_1}$ is the time-vertex based graph adjacency matrix with $I_1$ time-steps represented as graph nodes, and the weight matrix $\textbf{W}^{(r)} \in \mathbb{R}^{J_1 \times J_1}$ models information propagation between successive time-steps over $J_1$ features.

\textbf{The fRGTN Model}
The fast RGTN (fRGTN) is defined by approximating $\textbf{W}^{(r)} \approx \textbf{I}$ in gRGTN, which leads to a reduced forward pass, $\textbf{Y} = \sigma(\mathbf{R} \times_{2}^{2} \textbf{W}^{(x)} \times_2^1 \textbf{X})$, where $\mathbf{R} \in \mathbb{R}^{I_1 \times I_1}$ is a standard graph shift filter defined as $\mathbf{R} = (\textbf{I} + \textbf{A})$, as discussed in Section \ref{sec:GSP}. 

\textbf{The fRGTN-TT Model}
If the problem is inherently multi-modal, then the large dense layer matrices of the fRGTN can be tensorized and represented in the Tensor-Train format, as discussed in ~\citep{Novikov2015tnn}. This leads to the highly efficient fRGTN-TT model, which preserves the inherent multi-modality and has drastically lower parameter complexity.

\section{Multi-Graph Tensor Networks}
\label{sec:GTN}

\subsection{General Multi-Linear Graph Filter}

The time-based multi-linear graph filter, $\mathcalbf{R}$, was developed in ~\citep{xu2020recurrent} to model time-series problems through a time-graph adjacency matrix that reflects the temporal flow of information, as discussed in Section \ref{sec:RGTN}. For this filter to be extended to other domains, the underlying graph topology needs to be modified. More generally, given a weighted graph adjacency matrix, $\textbf{A} \in \mathbb{R}^{I_1 \times I_1}$, we can construct a multi-linear graph filter in the tensor domain, $\mathcalbf{F} \in \mathbb{R}^{J_1 \times I_1 \times J_1 \times I_1}$, as
\begin{equation}
\mathcalbf{F} = \tensorize \left(\textbf{I} + \beta \left(\textbf{A} \otimes \textbf{P} \right) \right)
\end{equation}
where the propagation matrix, $\textbf{P} \in \mathbb{R}^{J_1 \times J_1}$, models the flow of information between neighbouring vertices (as opposed to successive time-steps in the RGTN case), and $\beta$ is a learnable graph filter coefficient. This allows us to generalize the multi-linear graph filter $\mathcalbf{F}$ to any given graph domain of any data modality.

\begin{figure}[t]
	\centering
	\includegraphics[width=0.7\linewidth]{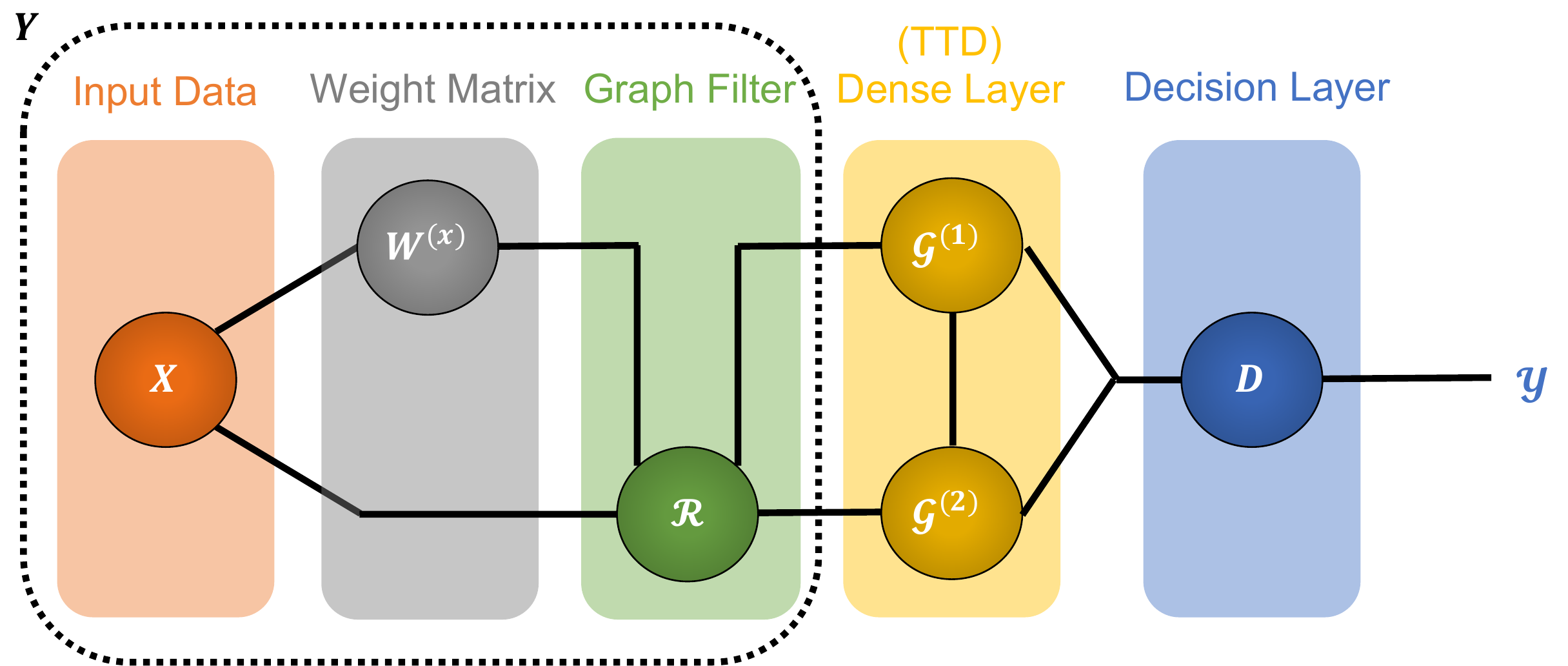}
	\caption{\textit{Illustration of the gRGTN model introduced in ~\citep{xu2020recurrent}. The section encircled in dotted line represents a general graph filtering operation for extracting hidden states in a time-series, $\textbf{Y}$, as discussed in Section \ref{sec:RGTN}.}}
	\label{fig:gGRTN}
\end{figure}
\begin{figure}[t]
	\centering
	\includegraphics[width=0.7\linewidth]{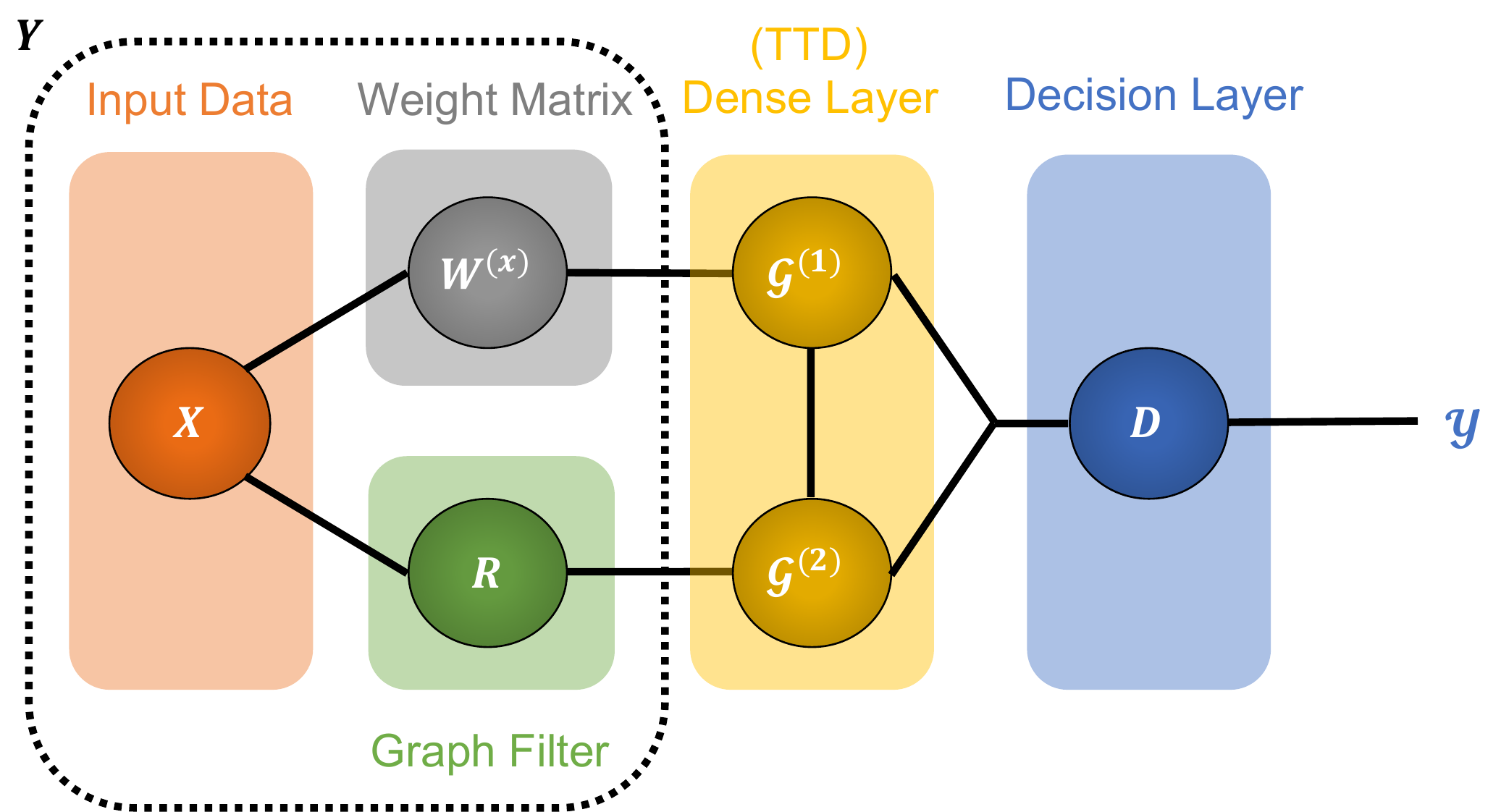}
	\caption{\textit{Illustration of the fRGTN model introduced in ~\citep{xu2020recurrent}. The section encircled in dotted line represents the fast graph filtering operation for extracting hidden states in a time-series, $\textbf{Y}$, as discussed in Section \ref{sec:RGTN}.}}
	\label{fig:sGRTN}
\end{figure}

\subsection{General Multi-Graph Tensor Network}
\label{general}
Consider a multi-graph learning problem where the input is an order-($M+1$) tensor $\mathcalbf{X} \in \mathbb{R}^{J_0 \times I_1 \times I_2 \times \cdots \times I_M}$ with $J_0$ features indexed along $M$ physical modes $\{I_1, I_2, \ldots, I_M\}$, such that a separate graph $\mathcal{G}^{(m)}$ is associated with each of the $I_m$ tensor modes, $m=1, \ldots, M$. For this problem, we define:
\begin{enumerate}
    \item $\mathcal{A} = \{ \textbf{A}^{(1)}, \textbf{A}^{(2)}, \ldots, \textbf{A}^{(M)}\}$, a set of adjacency matrices $\textbf{A}^{(m)} \in \mathbb{R}^{I_m \times I_m}$ constructed from the corresponding graphs $\mathcal{G}^{(m)}$.
    
    \item $\mathcal{W} = \{ \textbf{W}^{(1)}, \textbf{W}^{(2)}, \ldots, \textbf{W}^{(M)}\}$, a set of weight matrices $\textbf{W}^{(m)} \in \mathbb{R}^{J_m \times J_{m-1}}$ used for feature transforms, where $J_m$, for $m=1,\ldots,M$ controls the number of feature maps at every mode (dimension) $m$.
    
    \item $\mathcal{P} = \{ \textbf{P}^{(1)}, \textbf{P}^{(2)}, \ldots, \textbf{P}^{(M)}\}$, a set of propagation matrices $\textbf{P}^{(m)} \in \mathbb{R}^{J_m \times J_m}$, modelling the propagation of information over the neighbouring nodes of the graph $\mathcal{G}^{(m)}$.
    
    \item $\mathcal{B} = \{ \beta^{(1)}, \beta^{(1)}, \ldots, \beta^{(M)} \}$, a set of graph filtering coefficients, $\beta^{(m)} \in \mathbb{R}$, which scale the effects of the corresponding graph filters.
    
\end{enumerate}

Using the above domains and variables, we can now compute the multi-linear graph filters $\mathcalbf{F}^{(m)}$, as $\mathcalbf{F}^{(m)} = \tensorize(\textbf{I} + \beta^{(m)} (\textbf{A}^{(m)} \otimes \textbf{P}^{(m)}))$, for $m=1, \ldots, M$ graph domains. This allows us to define the \textit{general Multi-Graph Tensor Network} (gMGTN) layer with the forward pass defined as in Algorithm \ref{alg:MGTN}, which iterates the multi-linear graph filtering operation across all $M$ graph domains. The so defined forward pass generates a feature map, $\mathcalbf{Y} \in \mathbb{R}^{J_M \times I_1 \times \cdots \times I_M}$, from the input tensor, $\mathcalbf{X}$. Finally, non-linearity can be introduced through an activation function, $\sigma(\cdot)$.

\begin{algorithm}
\SetAlgoLined

    \textbf{Input:} $\mathcalbf{X}, \mathcal{A}, \mathcal{W}, \mathcal{P}, \sigma(\cdot)$
    
    \textbf{Output:} $\mathcalbf{Y}$
    
    \textbf{}\\

    $\mathcalbf{Y}$ = $\mathcalbf{X}$\\
    \For {m = 1, \ldots, M}{
 
    $\mathcalbf{F}^{(m)} = \tensorize(\textbf{I} + \beta^{(m)} (\textbf{A}^{(m)} \otimes \textbf{P}^{(m)}))$ 
  
    \textbf{update} $\mathcalbf{Y} = \mathcalbf{F}^{(m)} \times_{3,4}^{1, m+1} \textbf{W}^{(m)} \times_2^1 \mathcalbf{Y}$ 
 
    }
    $\mathcalbf{Y} = \sigma(\mathcalbf{Y})$
    \caption{gMGTN forward pass}\label{alg:MGTN}
\end{algorithm}

The above forward pass can also be written compactly through a series of tensor contractions as

\begin{equation}
\label{eq:gMGTN}
\mathcalbf{Y} = \sigma \big(\mathcalbf{F}^{(M)} \times_{3,4}^{1, M+1} \textbf{W}^{(M)} \times_2^1 \cdots \times_2^1 \mathcalbf{F}^{(1)} \times_{3,4}^{1, 2} \textbf{W}^{(1)} \times_2^1 \mathcalbf{X}\big)
\end{equation}

\subsection{Fast Multi-Graph Tensor Network}
\label{sec:sMGTN}

The gMGTN introduced above learns a propagation matrix, $\textbf{P}^{(m)}$, a weight matrix, $\textbf{W}^{(m)}$, and a graph filter coefficient, $\beta^{(m)}$, for each of the $M$ graphs within the gMGTN. For simplicity, let $J_m = J$ for $m=1,\ldots,M$; this results in a parameter complexity of $\mathcal{O}(MJ^2+M)$, which is linear in the number of graphs, $M$, and quadratic in the size of feature maps, $J$. This hinders the performance of gMGTN, since computation can become intractable for high dimensional multi-graph problems. To that end, we next develop the \textit{fast Multi-Graph Tensor Network} (fMGTN) as a low-complexity variant of the gMGTN.

Similar to ~\citep{xu2020recurrent}, we can reduce the parameter complexity of gMGTN by: (i) approximating $\textbf{P}^{(m)} \approx \textbf{I}$ for $m=1,\ldots,M$; and (ii) using one single weight matrix, $\textbf{W}^{(x)} \in \mathbb{R}^{J_1 \times J_0}$, for all of the graph domains, where $J_1$ controls the number of hidden units (feature maps). This allows us to compute the graph filters, $\mathbf{F}^{(m)}$, as $\mathbf{F}^{(m)} = (\textbf{I} + \beta^{(m)} \textbf{A}^{(m)})$, which leads to the fMGTN forward pass as described in Algorithm \ref{alg:sMGTN}.

\begin{algorithm}
\SetAlgoLined
 
    \textbf{Input:} $\mathcalbf{X}, \mathcal{A}, \mathbf{W}, \sigma(\cdot)$
    
    \textbf{Output:} $\mathcalbf{Y}$
    
    \textbf{}\\

    $\mathcalbf{Y} = \mathbf{W}^{(1)} \times_2^1 \mathcalbf{X}$\\
    \For {m = 1, \ldots, M}{

    $\mathbf{F}^{(m)} = \textbf{I} + \beta^{(m)} \textbf{A}^{(m)}$ 

    \textbf{update} $\mathcalbf{Y} = \mathbf{F} \times_{2}^{m} \mathcalbf{Y}$ 
    
    }
    
    $\mathcalbf{Y} = \sigma(\mathcalbf{Y})$ 
    
    \caption{fMGTN forward pass} \label{alg:sMGTN}
\end{algorithm}

The above forward pass can also be written compactly through a series of tensor contractions as
\begin{equation}
\label{eq:sMGTN}
\mathcalbf{Y} = \sigma \left(\mathbf{F}^{(M)} \times_{2}^{M+1} \cdots \times_2^4 \mathbf{F}^{(2)} \times_2^3 \mathbf{F}^{(1)} \times_2^2 \textbf{W}^{(1)} \times_2^1 \mathcalbf{X} \right)
\end{equation}

\begin{figure}[t!]
	\centering
	\includegraphics[width=0.7\linewidth]{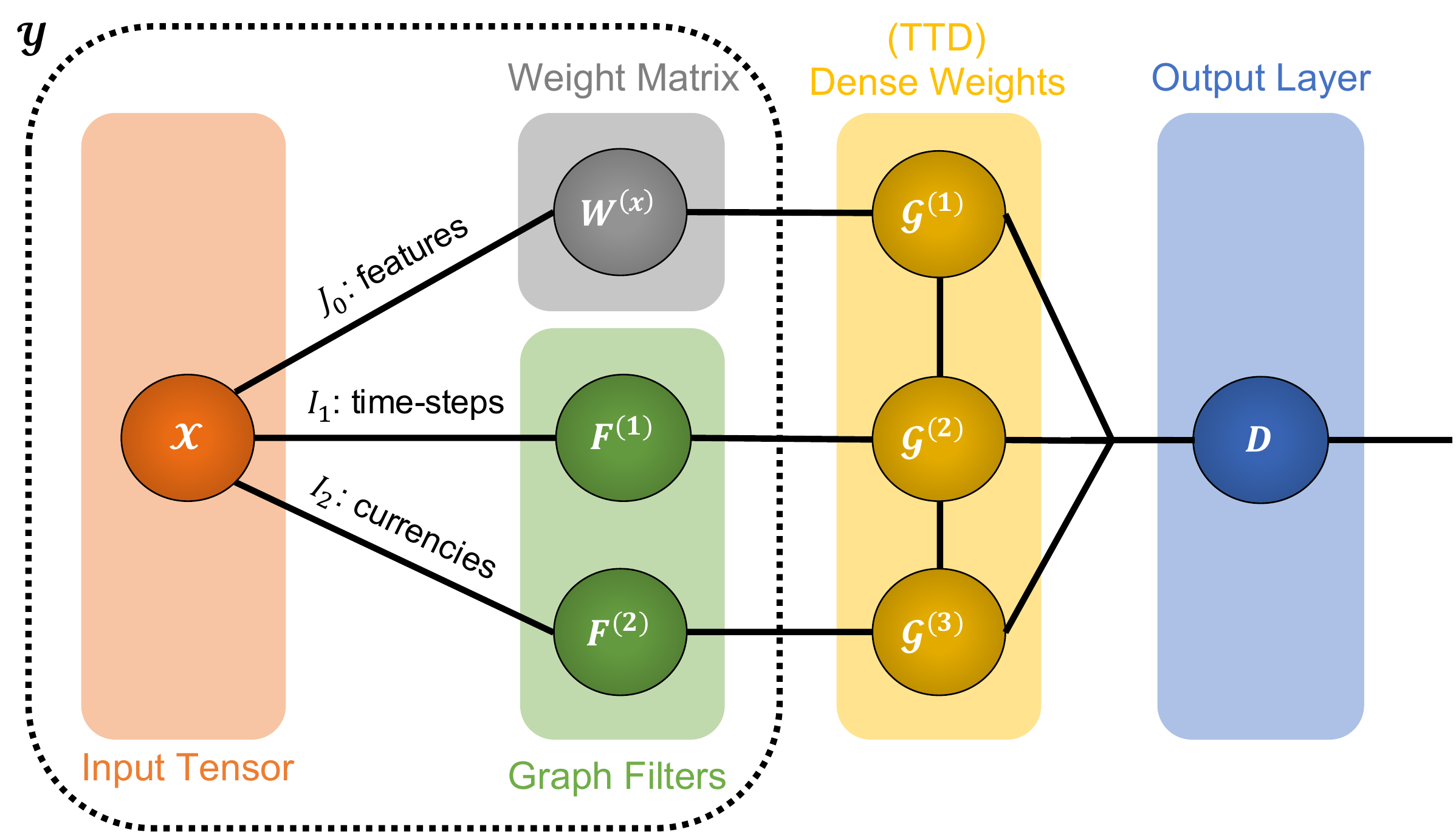}
	\caption{\textit{Tensor network representation of the fast Multi-Graph Tensor Network (fMGTN) used in the algorithmic trading experiment. The section encircled in dotted line denotes the multi-graph filtering operation for $M=2$ as in equation (\ref{eq:sMGTN}). The yellow region denotes a tensorized dense layer weight matrix, represented in the Tensor-Train format. The input data used for our experiment is an order-3 tensor with $J_0=4$ pricing features, $I_1=30$ past time-steps, and $I_2=9$ currencies, as discussed in Section \ref{sec:exp}. Note that we define a time-domain graph filter and a currency-domain graph filter for input data modes of respective dimensions $I_1$ and $I_2$.}}
	\label{fig:sMGTN}
\end{figure}

\begin{figure}[t!]
    \centering
    \includegraphics[width=1\textwidth]{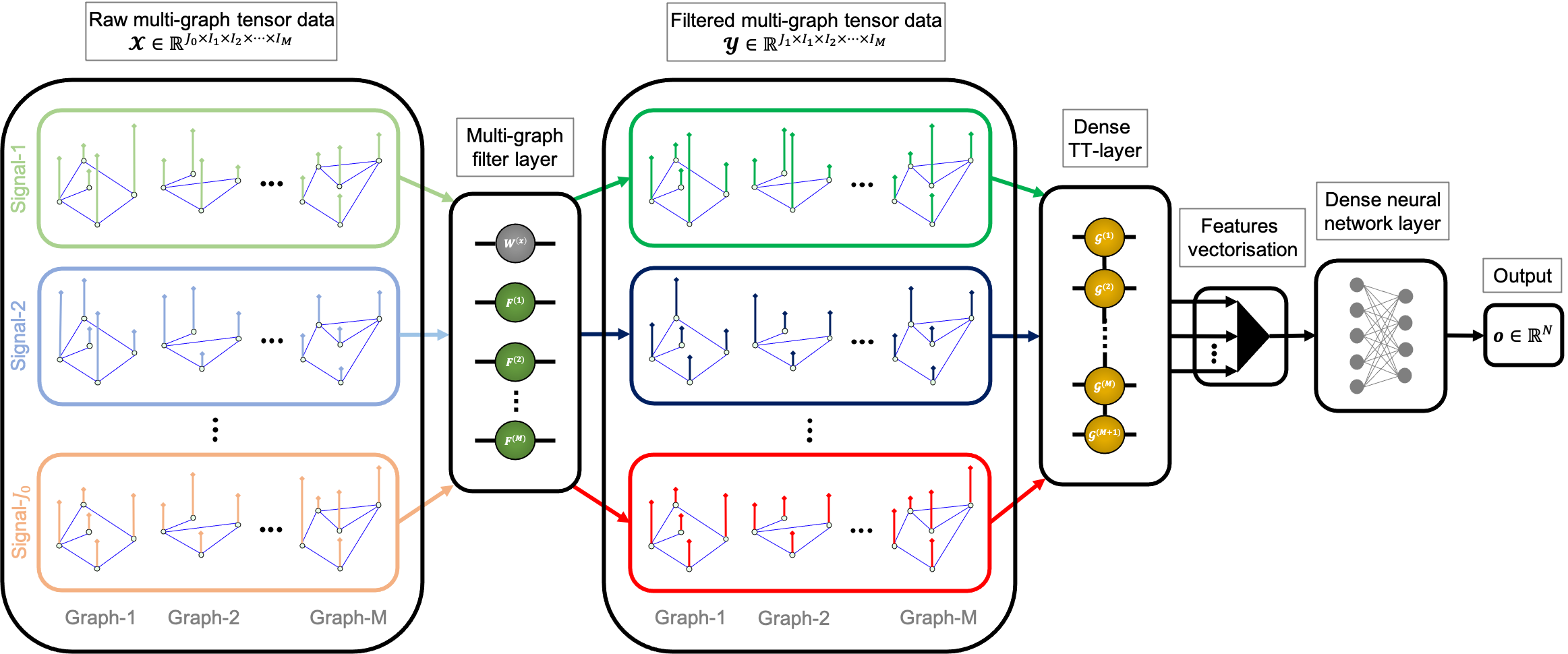}
    \caption{\textit{Principle of the proposed fMGTN framework, where the input data tensor, $\mathcalbf{X}$, is forward-passed from left-to-right to generate the overall output vector, $\mathbf{o}$. The model takes as input a tensor valued sample, $\mathcalbf{X}$, where $J_0$ different signals are indexed along $M$ different physical modes, with each physical mode being associated with a physical graph domain with $I_m$ nodes. The given input tensor is passed through the multi-graph filter layer (represented in tensor network notation), which generates a filtered representation of the signals while maintaining the underlying multi-graph and multi-dimensional structure. This multi-linear graph filtering constitutes a highly localized filtering operation, where locality is defined with respect to the topology of the graph. The filtered multi-graph tensor data is then passed through a dense layer in TTD format (represented in tensor network notation), which combines the localized features maps extracted previously via a global multi-linear map. The resulting global features are then vectorized (flattened) and passed through a final dense neural network layer to generate the desired output.}}
	\label{fig:mgtn_visualization}
\end{figure}

After extracting the feature map, $\mathcalbf{Y} \in \mathbb{R}^{J_1 \times I_1 \times \cdots \times I_M}$, it is customary to flatten the extracted features before passing them through dense layers of a neural network to generate the final output. To further reduce the parameter complexity, the weight matrices of the dense layers can also be tensorized and represented in the TT format, as discussed in ~\citep{Novikov2015tnn}. This further reduces the number of parameters, while maintaining compatibility with the inherent multi-modal nature of the problem. For clarity, an example of a fMGTN model which implements this series of contractions is shown in Figure \ref{fig:sMGTN}, using tensor network notation. In addition, Figure \ref{fig:mgtn_visualization} illustrates the same fMGTN architecture from a data processing point of view.

\subsection{Complexity Analysis}

Consider a multi-graph learning problem characterized by $M$ graphs, where each graph contains $I_1 = I_2 = \cdots = I_M = I$ nodes, and $J_0 = J_1 = \cdots = J_M = J$ features. In contrast to the gMGTN model, the proposed fMGTN does not need to learn $\textbf{P}^{(m)}$ or $\textbf{W}^{(m)}$, which reduces the parameter complexity of the forward pass to $\mathcal{O}(J^2 + M)$, but at the cost of lower expressive power. This is independent of graph dimensions, linear in the number of graphs, and quadratic in the feature dimensions. 

In comparison, standard matrix based methods would require a series of matricization operations to work with one single mode at time. For instance, when considering the time-mode $I_m$ only, it is necessary to reshape $\mathcalbf{X}$ along the corresponding modality and apply a RNN layer on the resulting matricized input, $\mathbf{X} \in \mathbb{R}^{I_m \times J_0 I_1 \cdots I_{m-1} I_{m+1} \cdots I_M}$. The matricization operation would incur an exponential increase in the feature dimensions, resulting in large weight matrices with parameter complexity of $\mathcal{O}(J^2 I^{2(M-1)})$, which can quickly become prohibitive for big data applications.

\section{Experiments}
\label{sec:exp}

To validate the expressive power of the MGTN model, we employed the proposed architecture in a number of experiments over domains of considerable variety, including algorithmic trading, bio-signal processing, climate change modelling, and air-quality forecasting. These experiments spam a diverse set of learning paradigms, including reinforcement learning, classification, and regression. Experimental results confirm the superiority of the proposed MGTN against a number of comparable deep learning architectures, both in terms of performance and parameter complexity.

\subsection{Algorithmic Trading}

We first explore the potential of the proposed MGTN model in the context of algorithmic trading, a notoriously difficult paradigm characterized by high-dimensional, multi-modal, noisy, and irregular data that pose significant challenges to traditional deep learning algorithms. 

\subsubsection{Financial Preliminaries}
\label{subsec:Carrygraph}

The FOREX market allows participants to trade pairs of currencies at a given \textit{spot rate}, which measures the value of a currency with respect to another currency at a given instant (e.g. EUR/USD spot rate of 1.2 implies that 1 Euro can be exchanged for 1.2 US Dollars). Alternatively, the participants can engage in forward contracts that allows them to exchange pairs of currencies on an agreed future date and at a specified \textit{forward rate}. If the forward rate of a given currency pair is higher than the current spot rate, then the numerator currency is expected to increase in value against the denominator currency and vice-versa.
There are many factors that can affect the movements of spot rates, although the most important factor is arguably the \textit{carry} factor: a tendency for high interest rate currencies to generate higher returns than the low interest rate ones. According to the \textit{interest-rate-parity} theory ~\citep{aliber1973interest}, the expectation of currency pairs moving in different directions, depending on the interest rate difference, is reflected in the difference between the spot rate and the forward rate. 
Therefore, for a pair of currencies, $i$ and $j$, we can construct a pairwise carry signal by computing $c_{i,j}=1-\frac{r_f}{r_s}$, where $r_f$ and $r_s$ denote respectively the forward rate and the spot rate of the currency pair. Finally, we can construct a carry graph adjacency matrix $\textbf{A}$ such that its entries, $a_{i,j}$, depend on the magnitude of the carry signal, $c_{i,j}$. Figure \ref{fig:carry_graph} (right) shows an example of the so constructed carry graph. 


The FOREX data are characterized by a number of properties that make classical machine learning techniques inadequate for their modelling; these include:
\begin{itemize}
    \item FOREX data are multi-modal in nature, since they contain multiple pricing information indexed over time and across several related assets, which results in large dimensional tensors whose computation suffers from the Curse of Dimensionality.
    \item Financial data is known to have low signal-to-noise ratio due to the arbitrage forces in the market ~\citep{de2020machine}, which makes training particularly susceptible to overfitting, especially for deep learning methods.
    \item Various market factors can influence the pricing at different degrees, depending on the time-horizon; this poses a multi-resolution problem that not many machine learning algorithms can handle.
\end{itemize}
The proposed MGTN is particularly suited to address the above challenges, as:
\begin{itemize}
    \item The multi-modal nature of FOREX data naturally leads to a tensor representation, which can be readily handled by the tensor network structure of our proposed model.
    \item The model can leverage the powerful low-rank compression and regularization properties of tensor networks, which are inherently immune to the Curse of Dimensionality and provide a regularization framework via TD that does not degrade the underlying data structure.
    \item Long-term market factors such as \textit{carry} can be encapsulated in graph filters that naturally allow for the pair-wise formulation of the FOREX data; this makes it possible to process high frequency pricing data through an economically meaningful low-frequency graph topology.
\end{itemize}

\subsubsection{Data Description}

\begin{figure}[h!]
	\centering
	\includegraphics[width=0.7\linewidth]{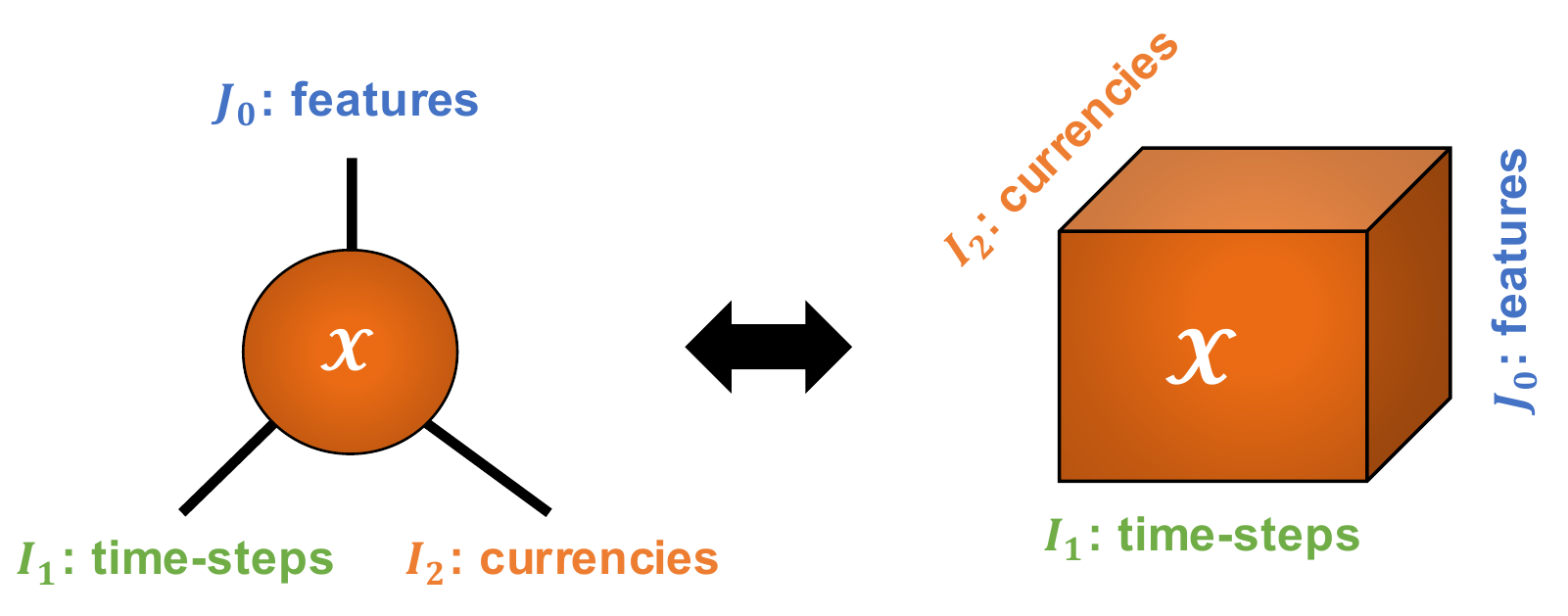}
	\caption{\textit{Input data tensor structure for the algorithmic trading experiment. The graph domain 1 corresponds to the time domain, while the graph domain 2 corresponds to the currency domain.}}
	\label{fig:trading_data_tensor}
\end{figure}

\begin{figure}[h!]
	\centering
	\includegraphics[width=0.7\linewidth]{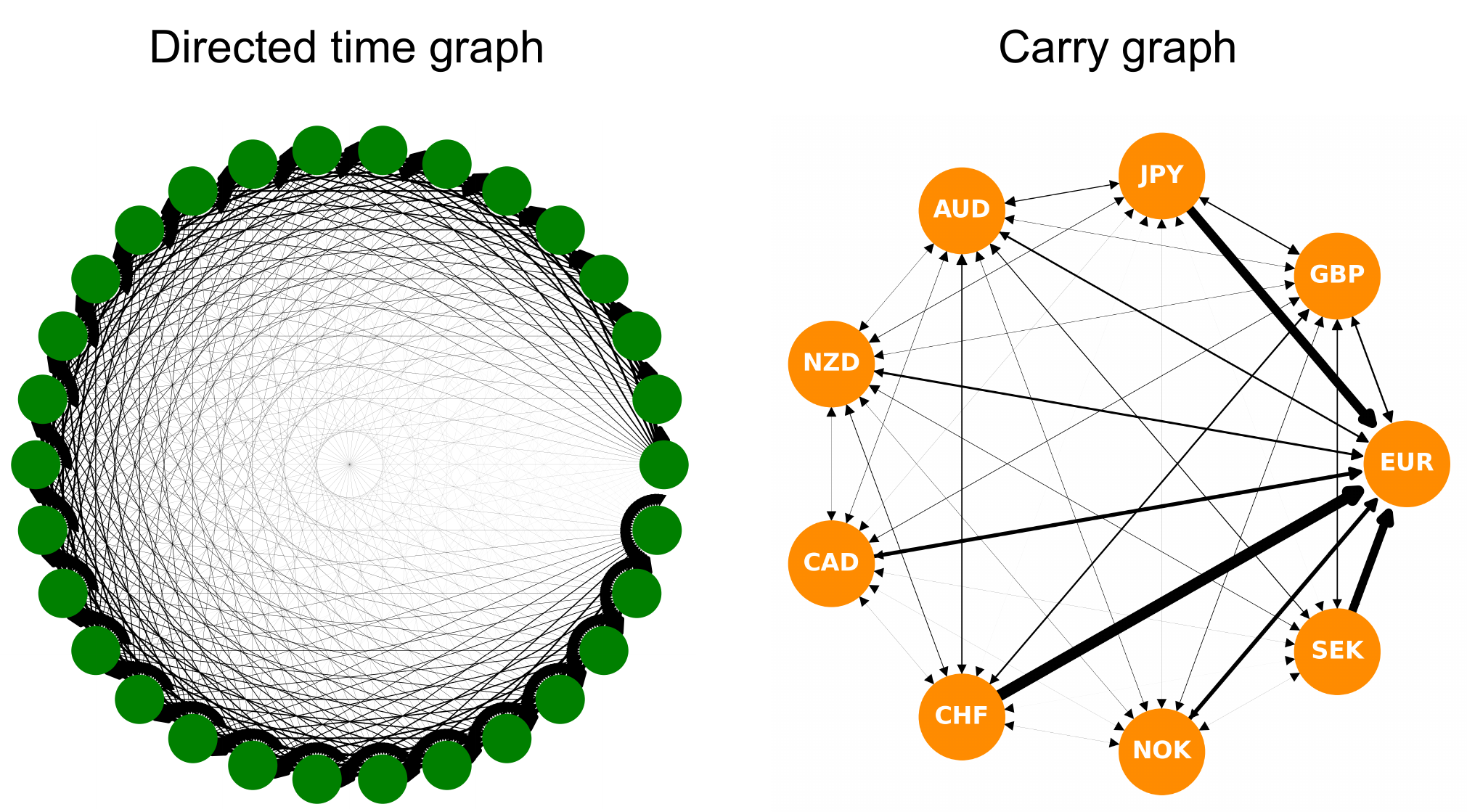}
	\caption{\textit{Graph domains for the algorithmic trading experiment. The time-domain graph is a directed graph where past states can influence future states but not vice-versa. The currency graph is also a directed graph, where the connection between currencies are proportional to the carry factor. In both graphs, thicker edges indicate stronger connection.}}
	\label{fig:carry_graph}
\end{figure}

Minute-wise spot-rate pricing data were used for the period between October 1\textsuperscript{st} 2019 and October 9\textsuperscript{th} 2019, for a total of 9 currencies. Training took place over the first 7 days, while out-of-sample performance evaluation was performed over the last 2 days. The considered features include Open/High/Low/Close (OHLC) spot-rates of the nine currencies, as shown in Figure \ref{fig:carry_graph}. 

We processed the data by computing the log-returns (log difference of spot-rates between successive time-steps) as $r_t = \ln(p_t)-\ln(p_{t-1})$. The log-returns were then aggregated into multi-modal input samples, $\mathcalbf{X} \in \mathbb{R}^{J_0 \times I_1 \times I_2}$, whereby log-returns were indexed along $J_0=4$ features (OHLC), $I_1=30$ past time-steps, and $I_2=9$ currencies. 
\subsubsection{Graph Domains}

For the given FOREX input data, we formulated the problem as a multi-graph learning problem, where each sample contains $J_0$ features indexed along $I_1$ time-steps and $I_2$ currencies, whereby a time-graph and a currency-graph are respectively associated with the time mode and the currency mode. More precisely, we formulated the time-graph for a total of $I_1=30$ time-steps, while the currency-graph was based on the carry-graph, as discussed in Section \ref{subsec:Carrygraph}. Finally, the respective graph filter for our fMGTN model was computed as discussed in Section \ref{sec:sMGTN}. This results in graph filters $\textbf{F}^{(1)}$ and $\textbf{F}^{(2)}$, as illustrated in Figure \ref{fig:sMGTN}.

\subsubsection{Models}

\begin{table}[h!]
    \small
    \centering
    \begin{tabular}{l l | l l l}
    
    \toprule
    
    \textbf{Model} & \textbf{Property} & \textbf{Layer 1} & \textbf{Layer 2} & \textbf{Layer 3} \\
        
    \midrule

    {fMGTN} & {Layer Type}   & {fMGTN}  & {TT-Dense}& {Dense}   \\ 
            & {Units}        & {16}     & {27}      & {2}      \\ 
            & {Activation}   & {relu}   & {relu}    & {linear}  \\ 
            & {TT-Rank}      & {n.a.}   & {(1,2,2,1)} & {n.a.}  \\ 
    
    \midrule
  
    {GRU}   & {Layer Type}   & {GRU}    & {Dense}   & {Dense}   \\ 
            & {Units}        & {16}     & {27}      & {2}      \\ 
            & {Activation}   & {relu}   & {relu}    & {linear}  \\ 
            & {TT-Rank}      & {n.a.}   & {n.a.}    & {n.a.}  \\ 
    
    \midrule
  
    {TTNN}  & {Layer Type}   & {TT-Dense}& {TT-Dense}& {Dense}   \\ 
            & {Units}        & {16}     & {27}      & {2}      \\ 
            & {Activation}   & {relu}   & {relu}    & {linear}  \\ 
            & {TT-Rank}      & {(1,2,2,1)}   & {(1,2,2,1)}    & {n.a.}  \\ 

    \midrule
  
    {GCN}   & {Layer Type}   & {GCN}    & {Dense}   & {Dense}   \\ 
            & {Units}        & {16}     & {27}      & {2}      \\ 
            & {Activation}   & {relu}   & {relu}    & {linear}  \\ 
            & {TT-Rank}      & {n.a.}   & {n.a.}    & {n.a.}  \\ 

    \bottomrule
  \end{tabular}
  \vspace{-1mm}
  \caption{\textit{Architecture of the models used in the experiment, based on: (i) fast Multi-Graph Tensor Network (MGTN), (ii) Gated Recurrent Unit (GRU) Recurrent Neural Network, (iii) Tensor-Train Neural Network (TTNN), and (iv) Graph Convolutional Network (GCN).}}
  \label{tab:mdl_architectures_trading}
\end{table}

To demonstrate the applicability and superiority of the proposed framework for the task of algorithmic trading, we used the fMGTN model as a feature extraction part of the deep Q network ~\citep{mnih2013playing} of a trading agent, and evaluated its performance against three commonly used agents based on: (i) Gated Recurrent Unit (GRU) Neural Network ~\citep{Chung2014}, (ii) Tensor-Train Neural Network (TTNN) ~\citep{Novikov2015tnn}, and (iii) Graph Convolutional Network (GCN) ~\citep{kipf2016semi}. Note that the multi-modal input data samples can be readily processed by the TTNN and the proposed fMGTN in their natural tensor form, as shown in Figure \ref{fig:sMGTN}. However, the input samples were matricized as $\mathbf{X} \in \mathbb{R}^{I_1 \times J_0I_2}$ and $\mathbf{X} \in \mathbb{R}^{I_2 \times J_0I_1}$ for compatibility with the GRU and the GCN agents, respectively.

For comparable results, the same model architecture specifications were applied across all agents, with the sole difference being the feature extraction method, as illustrated in Table \ref{tab:mdl_architectures_trading}. More specifically, each agent was based on a 3-layer architecture comprising: (i) a feature extraction layer with 16 units and ReLU activation, (ii) a dense layer with 27 units and ReLU activation, and (iii) a linear output layer with 2 units corresponding to buy and sell action values.  Each agent used a different feature extraction layer, which can be based on fMGTN, GRU, TTNN, or GCN. In addition, due to the inherent tensor representation, the dense layer following the fMGTN and TTNN layers was tensorized and represented in the TT format. All agents were trained using ADAM with a learning rate of $2\cdot10^{-4}$ and a mini-batch size of 64 for 15 episodes. Finally, the reward of the agents consisted of minute-wise log-returns. Our models were implemented\footnote{github.com/gylx/GTNRL-Trading} using TensorFlow 2.3.

\subsubsection{Experimental Results}

Four different financial metrics were used to assess the performance of the agents: \textit{Total return} (TR) measures the total percentage return generated by the agent in the episode; \textit{Sharpe ratio} (SR) measures the risk-adjusted return computed as $\frac{\mu_r}{\sigma_r}$, where $\mu_r$ is the average log-return and $\sigma_r$ the standard deviation of log-returns; \textit{Max Drawdown} (MD) measures the maximum percentage loss incurred by the agent during a consecutive period; \textit{Hit Rate} (HR) measures the percentage of profitable trades to total trades. Finally, we compared the parameter complexity by looking at the number of trainable parameters (NP). 

\begin{figure}[h!]
	\centering
	\includegraphics[width=0.9\linewidth]{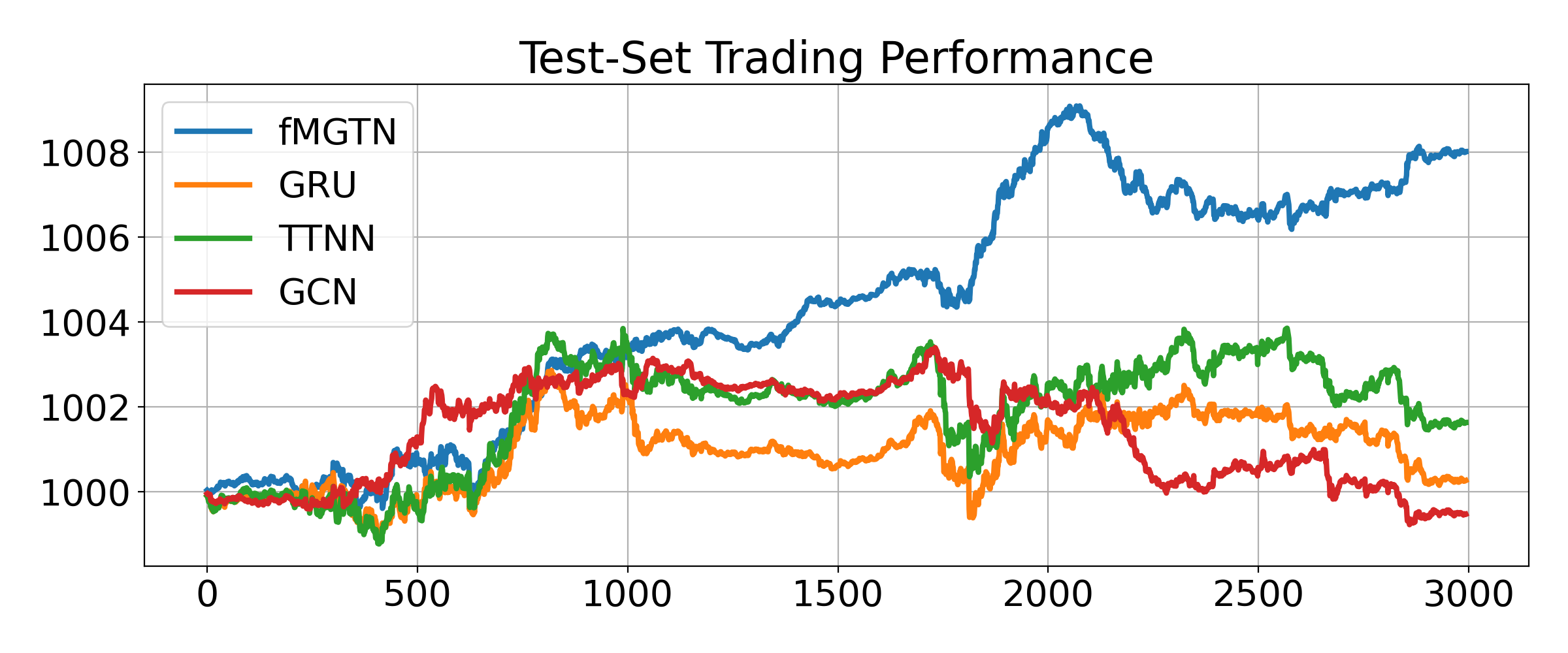}
	\caption{\textit{Out-of-sample trading performance of the considered agents, averaged over five European currencies. The vertical axis represents the investment growth of an initial portfolio value of 1000\$, while the horizontal axis represents time in minutes.}}
	\label{fig:performance}
\end{figure}

Highly superior performance for the fMGTN based agent was obtained across a basket of European currencies, both in terms of generated profits and other common financial metrics. As shown in Figure \ref{fig:performance}, the fMGTN agent generated substantial profit (0.8\%) during the out-of-sample testing period.

Table \ref{table:performance} summarizes the performance of the considered agents, with fMGTN significantly outperforming the other considered agents across a multitude of the most commonly used financial performance metrics. In addition, the fMGTN achieved the best performance at a drastically lower parameter complexity, using up to 90\% less trainable parameters compared to the GCN agent, and up to 80\% less compared to the GRU agent. 

\begin{table}[h!]
  \small
  \centering
  \begin{tabular}{l l l l l l}
  
    \toprule
     Agent & TR (\%) & SR & MD (\%) & HR(\%) & NP \\
    \midrule
  
  \textbf{fMGTN} & \textbf{0.8018} &  \textbf{0.0445} & \textbf{0.2893} & \textbf{52.8056} & \textbf{531} \\
  \midrule
  GRU & 0.0260 & 0.0012 & 0.3477 & 50.4008 & 3107 \\
  \midrule
  TTNN & 0.1628 & 0.0064 & 0.3493 & 50.6346 & {451} \\
  \midrule
  GCN & -0.0538 & -0.0032 & 0.4180 & 50.2338 & 5891 \\
    \bottomrule
  \end{tabular}
  \vspace{-1mm}
  \caption{\textit{Performance comparison for the considered agents for the task of algorithmic trading of currencies}}
  \label{table:performance}
\end{table}


\subsection{EEG Classification}

\begin{figure}[h!]
	\centering
	\includegraphics[width=0.7\linewidth]{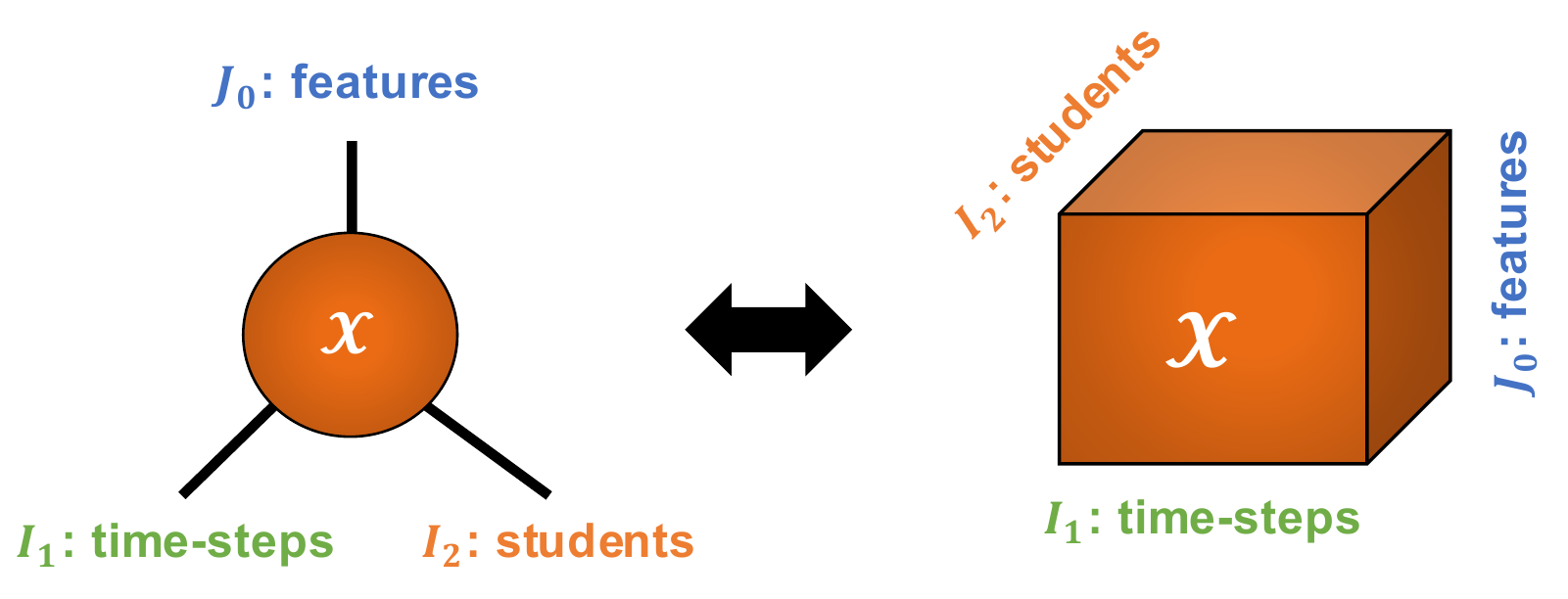}
	\caption{\textit{Tensor structure of the input data in the EEG classification experiment. For the given input tensor data, the graph domain 1 corresponds to the time domain, while the graph domain 2 corresponds to the student domain.}}
	\label{fig:eeg_data_tensor}
\end{figure}

\begin{figure}[h!]
	\centering
	\includegraphics[width=0.7\linewidth]{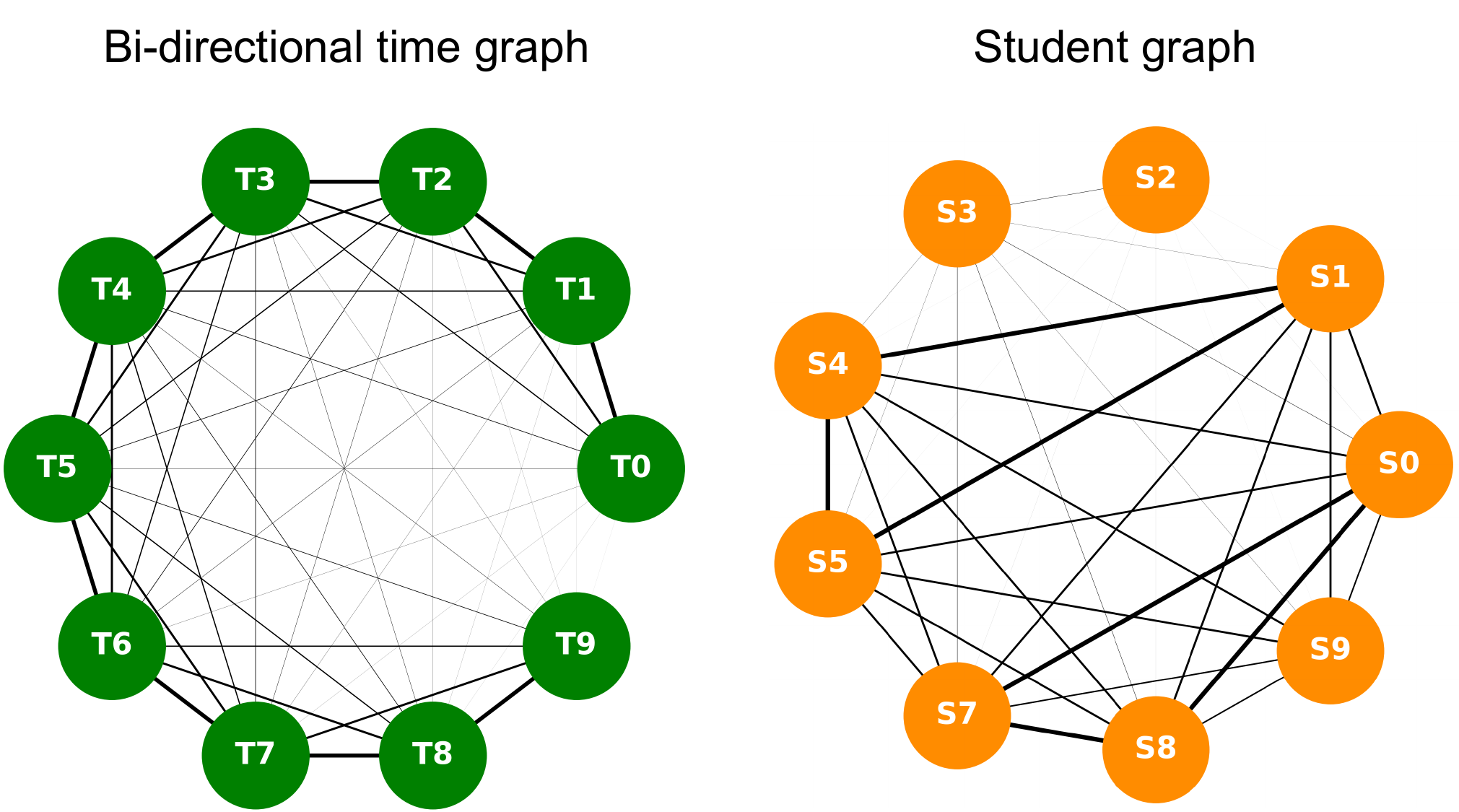}
	\caption{\textit{Graph domains for the EEG classification experiment. The time domain graph (left) is an undirected graph that captures the bi-directional flow of time, necessary for this experiment, and is constructed as discussed in Section \ref{subsubsec:graphdomains_eeg}. The student domain graph (right) is also an undirected graph where the connections are proportional to the demographic similarities between pairs of students. In both graph illustrations, thicker edges indicate stronger connections.}}
	\label{fig:data_tensor_structure_eeg}
\end{figure}

The task of this experiment is to classify the mental state of students (confused or not) from their electroencephalogram (EEG) readings as they watch online education videos ~\citep{wang2013using}. 

\subsubsection{Data Description}

The experiment data consists of 11 EEG time-series features (Attention, Meditation, Raw, Delta, Theta, Alpha1, Alpha2, Beta1, Beta2, Gamma1, Gamma2) recorded from each of the 9 students over 10 different videos. For the given dataset, we can maintain the inherent multi-modal structure of data by formulating the input samples  as tensors of order 3, $\mathcalbf{X} \in \mathbb{R}^{11 \times 10 \times 9}$, such that each sample contains $J_0 = 11$ features for each of the $I_1 = 10$ time-steps and $I_2 = 9$ students.

\subsubsection{Graph Domains}
\label{subsubsec:graphdomains_eeg}

To generalize the given problem as a multi-graph learning problem, we defined a time-graph and a student-graph associated with the time-mode and the student-mode, respectively. More precisely, we formulated a bi-directional time-graph adjacency matrix, $\textbf{A}^{(1)} \in \mathbb{R}^{10 \times 10}$, such that $\textbf{A}^{(1)} = \textbf{A}_t + \textbf{A}^T_t$, where $\textbf{A}_t$ is the triangular time-graph adjacency matrix as defined in ~\citep{xu2020recurrent} for a total of $I_1=10$ time-steps. For the student mode, we computed the student graph adjacency matrix $\textbf{A}^{(2)} \in \mathbb{R}^{9 \times 9}$, where the edge weights were computed via a Gaussian kernel, $A_{ij} = \text{exp}(-\frac{d(\textbf{s}_i, \textbf{s}_j)^2}{2 \sigma^2})$, with $\textbf{s}_i$ and $\textbf{s}_j$ as the vectors containing demographics information of the $i$-th and $j$-th student, $\sigma$ as a scaling factor, and $d(\cdot)$ as the function computing the Euclidean distance between the two vectors. 

\subsubsection{Models}

For comparable results, the proposed fMGTN model was implemented and compared to the same set of deep learning architectures as in Table \ref{tab:mdl_architectures_trading} from the algorithmic trading experiment. However, the number of hidden units and activation functions used across the three layers were changed to $(8, 27, 9)$ and $(\textit{tanh}, \textit{tanh}, \textit{linear})$ respectively, across all considered models. Finally, all models were trained with the same settings, that is using: (i) a RMS prop optimizer with a learning rate of $10^{-2}$ ~\citep{xu2021convergence}; (ii) a mean-squared-error loss function; (iii) a batch size of 32; and (iv) a total of 100 epochs. The first 70\% of the data was used for training purposes (20\% of which is used for validation), and the remaining 30\% for testing. 

Note that, due to their inherent tensor structure, the data samples were kept in their natural multi-modal form for the fMGTN and TTNN model, while their subsequent dense layer was represented in the TT format. For the GRU and GCN models, each input data sample was matricized to $\textbf{X} \in \mathbb{R}^{10 \times 99}$ and $\textbf{X} \in \mathbb{R}^{9 \times 110}$ along the time-mode and the student-mode, respectively.

\subsubsection{Experimental Results}

Experimental results are summarised in Table \ref{table:performance_eeg}, including the performance metrics such as the training accuracy (TRA) and testing accuracy (TEA), as well as complexity metrics such as the total number of trainable parameters (NP). The proposed fMGTN model outperformed all other considered models for the given task of EEG time-series classification, resulting in the highest accuracy score in out-of-sample testing, while using only a fraction of trainable parameters compared to other deep learning models, thus demonstrating promising results in terms of time-series classification.

\begin{table} [h!]
    \small
    \centering
    \begin{tabular}{l l l l l l}

        \toprule
        Model & TRA (\%) & TEA (\%) & NP \\
        \midrule
     
        \textbf{fMGTN} & \textbf{93.81} &  \textbf{56.10} & \textbf{585} \\
        \midrule
        GRU & 91.80 & 52.87 & 5055\\
        \midrule
        TTNN & 92.53 & 53.20 & 491\\
        \midrule
        GCN & 95.07 & 46.30 & 3103\\
        \bottomrule

    \end{tabular}
    \caption{\textit{Performance comparison for the considered models for the EEG classification experiment, measured in Training Accuracy (TRA), Testing Accuracy (TEA), and Number of Parameters (NP).}}
    \label{table:performance_eeg}
\end{table}

\subsection{Temperature Forecasting}

The task here was to forecast the monthly temperature levels across 92 different cities in the United States, by using the Climate Change: Earth Surface Temperature Data dataset ~\citep{rohde2013new}. 

\subsubsection{Data Description}

\begin{figure}[t!]
	\centering
	\includegraphics[width=0.7\linewidth]{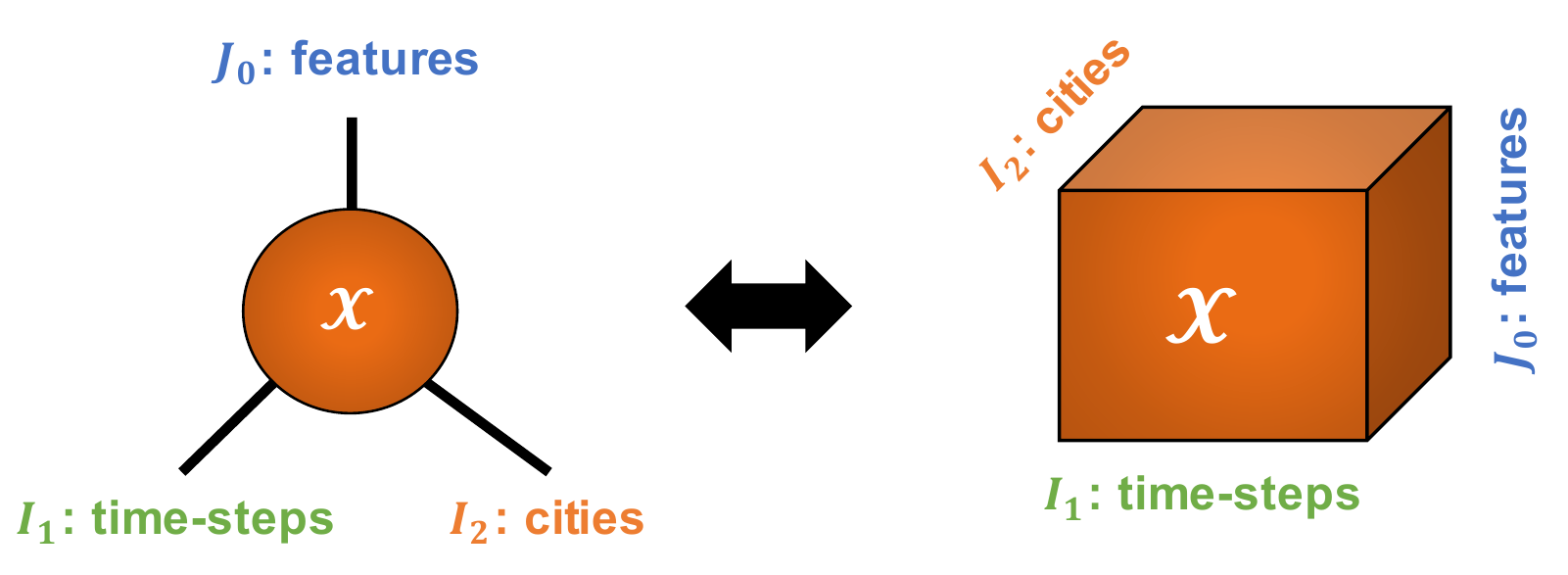}
	\caption{\textit{Input data tensor structure for the temperature forecasting experiment. The input tensor samples have $J_0$ features that live respectively on the time-domain and the city-domain.}}
	\label{fig:temp_data_tensor}
\end{figure}

\begin{figure}[t!]
	\centering
	\includegraphics[width=0.7\linewidth]{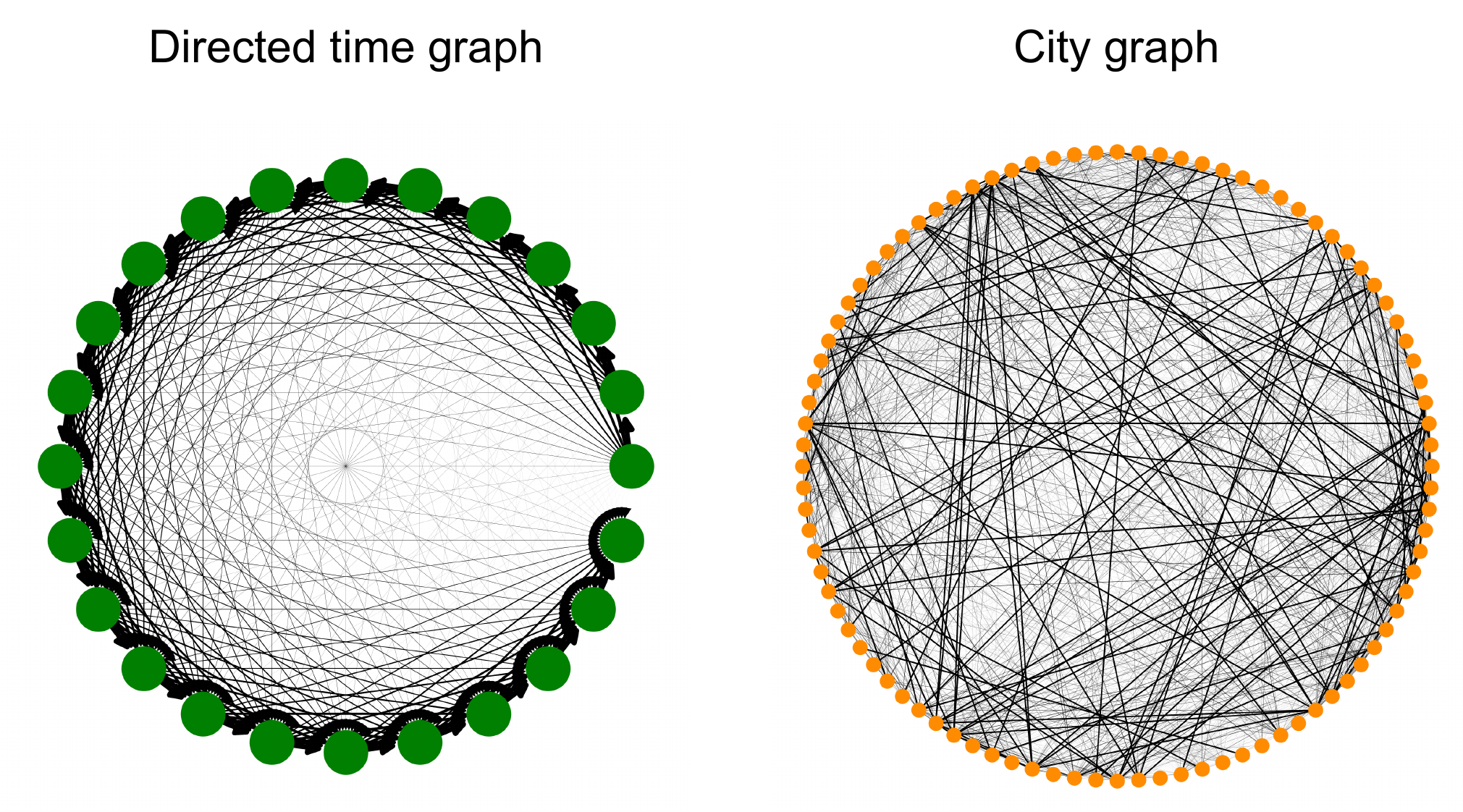}
	\caption{\textit{Graph domains for the temperature forecasting experiment. The time graph is a directed graph capturing the directed flow of time, while the city graph is an undirected graph where the connections between cities are proportional to their geographical distances. For both graphs, the strength of connections is proportional to the thickness of the edges.}}
	\label{fig:data_tensor_structure_temp}
\end{figure}

The experiment data consists of monthly recordings of (i) the average temperature value and (ii) the average temperature uncertainty, across multiple cities in the world. For our experiment, we focused on the 92 cities in the United States with the most complete set of data points. In addition to the two features provided in the dataset, we also included the sinusoidal encoding of temporal information as a third feature to account for potential seasonalities in the time-series. This led to an order-3 tensor representation of the input data, $\mathcalbf{X} \in \mathbb{R}^{3 \times 24 \times 92}$, such that each sample contains $J_0 = 3$ features for each of the $I_1 = 24$ time-steps and $I_2 = 92$ cities.

\subsubsection{Graph Domains}

To formulate the given problem as a multi-graph learning problem, we defined a time-graph and a city-graph associated with the time-mode and the city-mode respectively. More precisely, we formulated the directed time-graph, $\textbf{A}^{(1)} \in \mathbb{R}^{24 \times 24}$, in the same way as in ~\citep{xu2020recurrent} for a total of $I_1=24$ time-steps (2 years). For the city-graph, we computed the city graph adjacency matrix, $\textbf{A}^{(2)} \in \mathbb{R}^{92 \times 92}$, such that the edge weights were calculated from a Gaussian kernel applied on the latitude and longitude information of different cities.

\subsubsection{Models}


For comparable results, the proposed fMGTN model was implemented and compared against the same set of deep learning architectures as indicated in Table \ref{tab:mdl_architectures_trading} for the algorithmic trading experiment. For the present experiment, however, the number of hidden units and activation functions used across the three layers were changed to $(32, 8, 92)$ and $(\textit{tanh}, \textit{tanh}, \textit{linear})$ respectively, for all considered models. Finally, all models were trained with the same settings, that is using: (i) a RMS prop optimizer with a learning rate of $10^{-2}$; (ii) a mean-squared-error loss function; (iii) a batch size of 32; and (iv) a total of 30 epochs. The first 70\% of the data was used for training purposes (20\% of which was used for validation), and the remaining 30\% for testing. 

Note that, due to their inherent tensor structure, the data samples were kept in their natural multi-modal form for the fMGTN and TTNN model, while their subsequent dense layer was represented in the TT format. For the GRU and GCN model, each input data sample was instead matricized to $\textbf{X} \in \mathbb{R}^{24 \times 276}$ and $\textbf{X} \in \mathbb{R}^{92 \times 72}$ along the time-mode and the city-mode, respectively.

\subsubsection{Experimental Results}

Experimental results are summarised in Table \ref{table:performance_temperature}, including performance metrics such as the training mean-square-error (TRMSE) and testing mean-squared-error (TEMSE), as well as the complexity metrics such as the total number of trainable parameters (NP). The proposed fMGTN model is shown to out-perform all other considered models for the task of temperature forecasting, resulting in the best out-of-sample mean-squared-error, while using only a fraction of trainable parameters compared to other deep learning models, thus fully demonstrating the potential of the proposed framework in regression tasks as well.

\begin{table} [h!]
    \small
    \centering
    \begin{tabular}{l l l l l l}

        \toprule
        Model & TRMSE (\%) & TEMSE (\%) & NP \\
        \midrule
     
        \textbf{fMGTN} & \textbf{0.0206} &  \textbf{0.0186} & \textbf{1894} \\
        \midrule
        GRU & 0.0237 & 0.0226 & 36740\\
        \midrule
        TTNN & 0.0237 & 0.0188 & 2600\\
        \midrule
        GCN & 0.0390 & 0.0305 & 84395\\
        \bottomrule

    \end{tabular}
    \caption{\textit{Performance comparison for the considered models for the temperature forecasting task, measured in Training Mean-Squared-Error (TRMSE), Testing Mean-Squared Error (TEMSE), and Number of Parameters (NP).}}
    \label{table:performance_temperature}
\end{table}

\subsection{Air Quality Forecasting}

\begin{figure}[h!]
	\centering
	\includegraphics[width=0.7\linewidth]{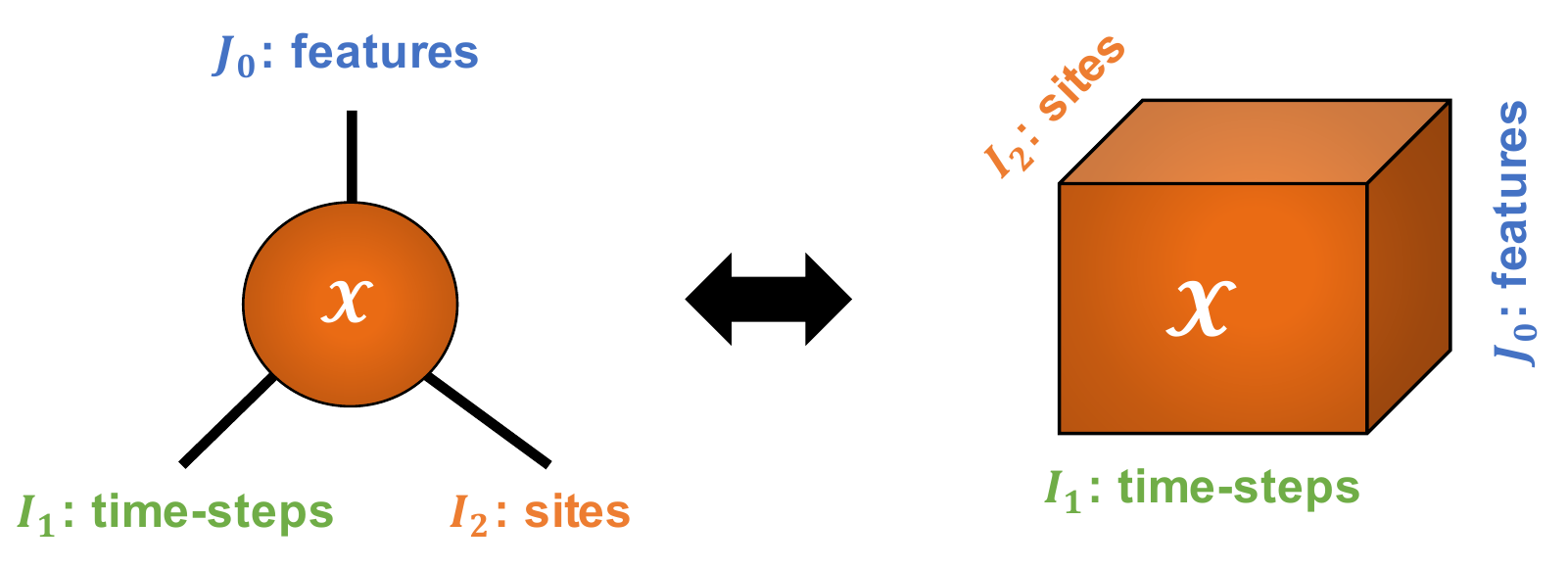}
	\caption{\textit{Tensor structure of the input data for the air quality forecasting experiment. The input tensor contains features that live simultaneously on the time-domain and the site-domain, both being irregular domains that can be captured through graphs.}}
	\label{fig:air_data_tensor}
\end{figure}

\begin{figure}[h!]
	\centering
	\includegraphics[width=0.7\linewidth]{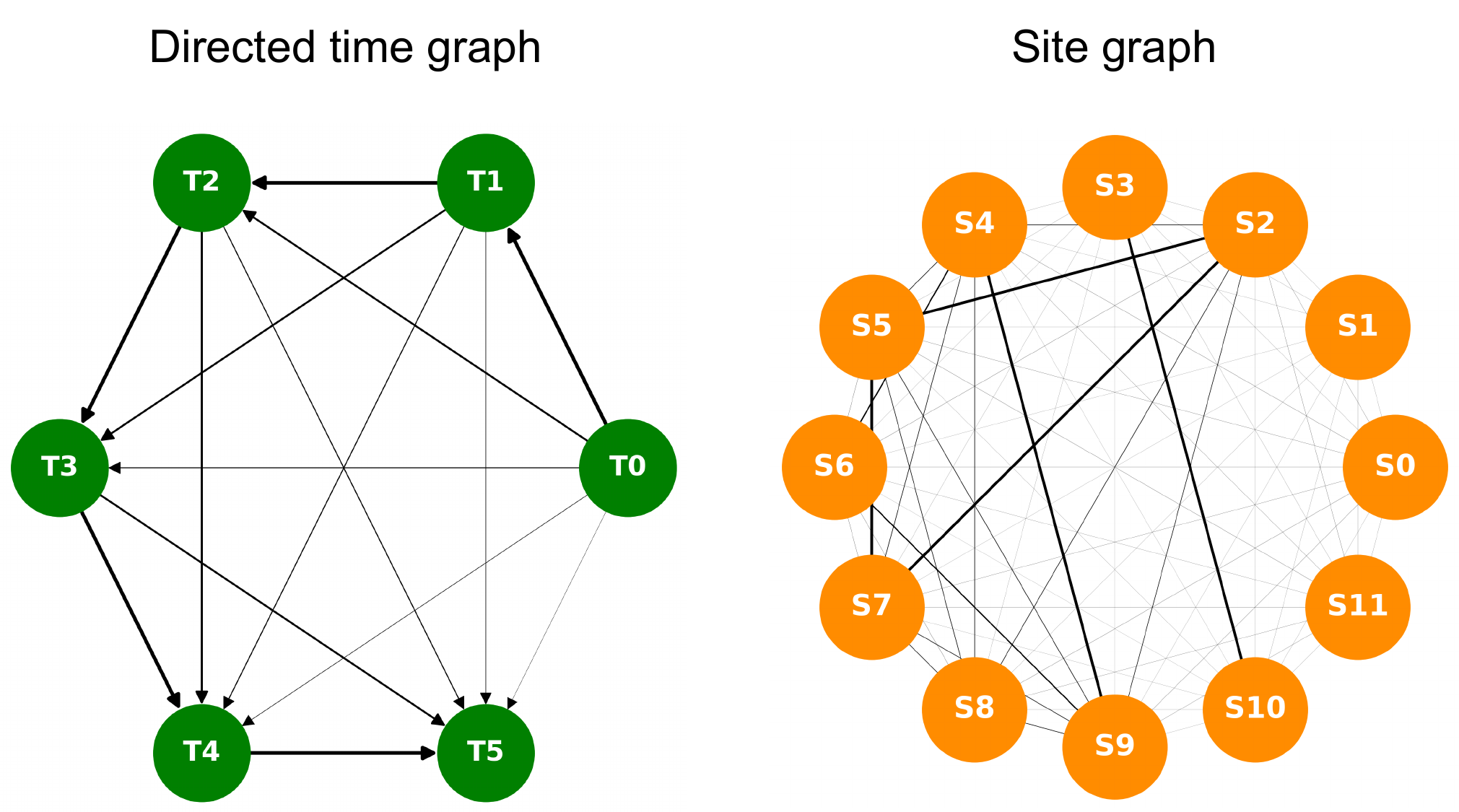}
	\caption{\textit{Graph domains for the air quality forecasting experiment. The time graph is a directed graph that captures the directed flow of information over time, while the site graph is a undirected graph where the connection between sites are proportional to the correlations of measured data between the pairs of sites, which are assumed to be proportional to their geographical distances.}}
	\label{fig:data_tensor_structure_air}
\end{figure}

The learning task here was to forecast the future PM2.5 level across 12 different regions in China, using the Beijing Multi-Site Air-Quality dataset ~\citep{zhang2017cautionary}. 

\subsubsection{Data Description}

The experiment data consists of 12 time-series features (PM2.5, PM10, SO2, NO2, CO, O3, TEMP, PRES, DEWP, RAIN, wd, WSPM) recorded hourly across 12 different geographical sites in China. For the experiment, the given raw time-series features were pre-processed through suitable one-hot-encoding of categorical variables (such as the window direction wd), leading to a total of 27 features. The multi-site nature of the experiment data naturally admits a tensor representation, leading to sample tensors of order 3, $\mathcalbf{X} \in \mathbb{R}^{27 \times 6 \times 12}$, such that each sample contains $J_0 = 27$ features for each of the $I_1 = 6$ time-steps and $I_2 = 12$ geographical sites.

\subsubsection{Graph Domains}

To formulate the given problem as a multi-graph learning problem, we defined a time-graph and a site-graph associated with the time-mode and the site-mode respectively. More precisely, we formulated the directed time-graph, $\textbf{A}^{(1)} \in \mathbb{R}^{6 \times 6}$, in the same way as in ~\citep{xu2020recurrent} for a total of $I_1=6$ time-steps. For the site-graph, we computed the graph adjacency matrix, $\textbf{A}^{(2)} \in \mathbb{R}^{12 \times 12}$, from the correlation values of the recorded data across different sites, assuming that sites with closer geographical proximity exhibit more correlated data recordings.

\subsubsection{Models}


The set of neural network architectures implemented for this experiment was the same as those indicated in Table \ref{tab:mdl_architectures_trading} for the algorithmic trading experiment. However, the number of hidden units and activation functions used across the three layers was changed to $(8, 8, 12)$ and $(\textit{tanh}, \textit{tanh}, \textit{linear})$ respectively, for all considered models. Finally, all models were trained with the same settings, that is using: (i) RMS prop optimizer with a learning rate of $20^{-3}$; (ii) mean-squared-error loss function; (iii) batch size of 32; and (iv) a total of 50 epochs. The first 70\% of the data was used for training purposes (20\% of which was used for validation), and the remaining 30\% for testing. 

Note that, due to their inherent tensor structure, the data samples were kept in their natural multi-modal form for the fMGTN and TTNN model, while their subsequent dense layer was represented in the TT format. For the GRU and GCN models, each input data sample was instead matricized to $\textbf{X} \in \mathbb{R}^{6 \times 324}$ and $\textbf{X} \in \mathbb{R}^{12 \times 162}$ along the time-mode and the site-mode, respectively.

\subsubsection{Experimental Results}

Experimental results are summarised in Table \ref{table:performance_air}. The proposed fMGTN model is shown to out-perform all other considered models for the given task of air quality forecasting, resulting in the best out-of-sample mean-squared-error.

\begin{table} [h!]
    \small
    \centering
    \begin{tabular}{l l l l l l}

        \toprule
        Model & TRMSE (\%) & TEMSE (\%) & NP \\
        \midrule
     
        \textbf{fMGTN} & \textbf{0.0936} &  \textbf{0.1348} & \textbf{486} \\
        \midrule
        GRU & 0.1197 & 0.1903 & 8516 \\
        \midrule
        TTNN & 0.1204 & 0.1542 & 384\\
        \midrule
        GCN & 0.1400 & 0.2036 & 1296\\
        \bottomrule

    \end{tabular}
    \caption{\textit{Performance comparison for the considered models for the task of air quality forecasting, measured in Training Mean-Squared-Error (TRMSE), Testing Mean-Squared Error (TEMSE), and Number of Parameters (NP).}}
    \label{table:performance_air}
\end{table}

\section{Conclusion}
We have introduced a novel deep learning framework which is suitable for multimodal data acquired on irregular domains. This has been achieved by leveraging on the virtues of graphs and tensors to provide an efficient modelling strategy in a deep learning setting, and across diverse learning paradigms, including regression, classification and reinforcement learning. The resulting \textit{Multi-Graph-Tensor-Network} (MGTN) has been shown to be capable of handling irregular data residing on multiple graph domains, while simultaneously benefiting from the compression properties of tensor networks to enhance the modelling power and drastically reduce parameter complexity. Experimental results have validated the potential of the proposed framework for integrating graphs, tensors, and neural networks, through performance records which are superior to individual performances in any of the three constituent domains.

Future research directions include leveraging on the versatility of the proposed framework to investigate its potential in numerous applications that share the modelling setup considered here. For example, the graph filters within the MGTN allow for the modelling of irregular data defined on one or multiple graph domains, a typical setting in social networks, recommender systems, and traffic forecasting. In addition, the tensor network structure of the MGTN allows for the modelling of high-dimensional data at a low complexity, which appeals to problems including multi-sensor processing, video classification, and natural language processing. Future research on spectral MGTN models can also potentially improve the modelling power of MGTN through spectral graph filtering techniques.


\acks{Y.L.X. is supported by
an EPSRC Departmental Scholarship. K.K. is
supported by an EPSRC International Doctoral Scholarship.}






\vskip 0.2in
\bibliography{sample}

\begin{thebibliography}{26}
\providecommand{\natexlab}[1]{#1}
\providecommand{\url}[1]{\texttt{#1}}
\expandafter\ifx\csname urlstyle\endcsname\relax
  \providecommand{\doi}[1]{doi: #1}\else
  \providecommand{\doi}{doi: \begingroup \urlstyle{rm}\Url}\fi

\bibitem[Aliber(1973)]{aliber1973interest}
R.~Z. Aliber.
\newblock The interest rate parity theorem: A reinterpretation.
\newblock \emph{Journal of Political Economy}, 81\penalty0 (6):\penalty0
  1451--1459, 1973.

\bibitem[Chung et~al.(2014)Chung, Gulcehre, Cho, and Bengio]{Chung2014}
J.~Chung, C.~Gulcehre, K.~Cho, and Y.~Bengio.
\newblock Empirical evaluation of gated recurrent neural networks on sequence
  modeling.
\newblock In \emph{Proceedings of the NIPS 2014 Workshop on Deep Learning},
  December 2014.

\bibitem[{Cichocki}(2014)]{Cichocki2014}
A.~{Cichocki}.
\newblock Era of big data processing: A new approach via tensor networks and
  tensor decompositions.
\newblock In \emph{Proceedings of the International Workshop on Smart
  Info-Media Systems in Asia}, March 2014.

\bibitem[{Cichocki} et~al.(2015){Cichocki}, {Mandic}, {De Lathauwer}, {Zhou},
  {Zhao}, {Caiafa}, and {PHAN}]{7038247}
A.~{Cichocki}, D.~{Mandic}, L.~{De Lathauwer}, G.~{Zhou}, Q.~{Zhao},
  C.~{Caiafa}, and H.~A. {PHAN}.
\newblock Tensor decompositions for signal processing applications: From
  two-way to multiway component analysis.
\newblock \emph{IEEE Signal Processing Magazine}, 32\penalty0 (2):\penalty0
  145--163, March 2015.

\bibitem[Cichocki et~al.(2016)Cichocki, Lee, Oseledets, Phan, Zhao, Mandic,
  et~al.]{cichocki2016tensor}
A.~Cichocki, N.~Lee, I.~Oseledets, A.~Phan, Q.~Zhao, D.~P. Mandic, et~al.
\newblock Tensor networks for dimensionality reduction and large-scale
  optimization. part 1: Low-rank tensor decompositions.
\newblock \emph{Foundations and Trends{\textregistered} in Machine Learning},
  9\penalty0 (4-5):\penalty0 249--429, 2016.

\bibitem[Cohen et~al.(2016)Cohen, Sharir, and Shashua]{cohen2016expressive}
N.~Cohen, O.~Sharir, and A.~Shashua.
\newblock On the expressive power of deep learning: A tensor analysis.
\newblock In \emph{Proceedings of the Conference on Learning Theory}, pages
  698--728, 2016.

\bibitem[de~Prado(2020)]{de2020machine}
M.~L. de~Prado.
\newblock \emph{Machine Learning for Asset Managers}.
\newblock Cambridge University Press, 2020.

\bibitem[Defferrard et~al.(2016)Defferrard, Bresson, and
  Vandergheynst]{NIPS2016_04df4d43}
M.~Defferrard, X.~Bresson, and P.~Vandergheynst.
\newblock Convolutional neural networks on graphs with fast localized spectral
  filtering.
\newblock In \emph{Proceedings of the Advances in Neural Information Processing
  Systems}, volume~29, pages 3844--3852, 2016.

\bibitem[Dolgov and Savostyanov(2014)]{Dolgov2014}
S.V. Dolgov and D.V. Savostyanov.
\newblock {Alternating minimal energy methods for linear systems in higher
  dimensions}.
\newblock \emph{SIAM Journal on Scientific Computing}, 36\penalty0
  (5):\penalty0 A2248--A2271, 2014.

\bibitem[Kipf and Welling(2017)]{kipf2016semi}
T.~N. Kipf and M.~Welling.
\newblock Semi-supervised classification with graph convolutional networks.
\newblock In \emph{Proceedings of the International Conference on Learning
  Representations (ICLR)}, 2017.

\bibitem[Li et~al.(2016)Li, Tarlow, Brockschmidt, and
  Zemel]{DBLP:journals/corr/LiTBZ15}
Y.~Li, D.~Tarlow, M.~Brockschmidt, and R.~S. Zemel.
\newblock Gated graph sequence neural networks.
\newblock In \emph{Proceedings of the 4th International Conference on Learning
  Representations}, 2016.

\bibitem[Mnih et~al.(2013)Mnih, Kavukcuoglu, Silver, Graves, Antonoglou,
  Wierstra, and Riedmiller]{mnih2013playing}
V.~Mnih, K.~Kavukcuoglu, D.~Silver, A.~Graves, I.~Antonoglou, D.~Wierstra, and
  M.~Riedmiller.
\newblock Playing {A}tari with deep reinforcement learning.
\newblock \emph{ArXiv e-prints}, December 2013.

\bibitem[Monti et~al.(2017)Monti, Bronstein, and Bresson]{NIPS2017_2eace51d}
F.~Monti, M.~Bronstein, and X.~Bresson.
\newblock Geometric matrix completion with recurrent multi-graph neural
  networks.
\newblock In \emph{Proceedings of the Advances in Neural Information Processing
  Systems}, volume~30, pages 3697--3707, 2017.

\bibitem[Novikov et~al.(2015)Novikov, Podoprikhin, Osokin, and
  Vetrov]{Novikov2015tnn}
A.~Novikov, D.~Podoprikhin, A.~Osokin, and D.~P. Vetrov.
\newblock Tensorizing neural networks.
\newblock In \emph{Proceedings of the Advances in Neural Information Processing
  Systems (NIPS)}, pages 442--450, 2015.

\bibitem[Oseledets(2011)]{oseledets2011tensor}
I.~V. Oseledets.
\newblock Tensor-train decomposition.
\newblock \emph{SIAM Journal on Scientific Computing}, 33\penalty0
  (5):\penalty0 2295--2317, 2011.

\bibitem[Rohde et~al.(2013)Rohde, Muller, Jacobsen, Muller, Perlmutter,
  Rosenfeld, Wurtele, Groom, and Wickham]{rohde2013new}
R.~Rohde, R.~A. Muller, R.~Jacobsen, E.~Muller, S.~Perlmutter, A.~Rosenfeld,
  J.~Wurtele, D.~Groom, and C.~Wickham.
\newblock A new estimate of the average earth surface land temperature spanning
  1753 to 2011.
\newblock \emph{Geoinfor Geostat}, 7:\penalty0 2, 2013.

\bibitem[Shuman et~al.(2013)Shuman, Narang, Frossard, Ortega, and
  Vandergheynst]{shuman2013emerging}
D.~I. Shuman, S.~K. Narang, P.~Frossard, A.~Ortega, and P.~Vandergheynst.
\newblock The emerging field of signal processing on graphs: Extending
  high-dimensional data analysis to networks and other irregular domains.
\newblock \emph{IEEE {S}ignal {P}rocessing {M}agazine}, 30\penalty0
  (3):\penalty0 83--98, 2013.

\bibitem[Stankovic et~al.(2020{\natexlab{a}})Stankovic, Mandic, Dakovic,
  Brajovic, Scalzo, and Constantinides]{stankovic2019graphII}
L.~Stankovic, D.~Mandic, M.~Dakovic, M.~Brajovic, B.~Scalzo, and A.~G.
  Constantinides.
\newblock Data analytics on graphs. {P}art {II}: Signals on graphs.
\newblock \emph{Foundations and Trends in Machine Learning}, 13\penalty0
  (2--3):\penalty0 158--331, 2020{\natexlab{a}}.

\bibitem[Stankovic et~al.(2020{\natexlab{b}})Stankovic, Mandic, Dakovic,
  Brajovic, Scalzo, and Constantinides]{stankovic2019graph}
L.~Stankovic, D.~Mandic, M.~Dakovic, M.~Brajovic, B.~Scalzo, and
  T.~Constantinides.
\newblock Data analytics on graphs. {P}art {I}: Graphs and spectra on graphs.
\newblock \emph{Foundations and Trends in Machine Learning}, 13\penalty0
  (1):\penalty0 1--157, 2020{\natexlab{b}}.

\bibitem[Wang et~al.(2013)Wang, Li, Hu, Yang, Meng, and Chang]{wang2013using}
H.~Wang, Y.~Li, X.~Hu, Y.~Yang, Z.~Meng, and K.~M. Chang.
\newblock Using {EEG} to improve massive open online courses feedback
  interaction.
\newblock In \emph{Proceedings of AIED Workshops}, 2013.

\bibitem[Wu et~al.(2020)Wu, Pan, Chen, Long, Zhang, and
  Philip]{wu2020comprehensive}
Z.~Wu, S.~Pan, F.~Chen, G.~Long, C.~Zhang, and S.~Y. Philip.
\newblock A comprehensive survey on graph neural networks.
\newblock \emph{IEEE Transactions on Neural Networks and Learning Systems},
  pages 1--21, 2020.

\bibitem[Xu et~al.(2021)Xu, Zhang, Zhang, and Mandic]{xu2021convergence}
D.~Xu, S.~Zhang, H.~Zhang, and D.~P. Mandic.
\newblock Convergence of the {RMSP}rop deep learning method with penalty for
  nonconvex optimization.
\newblock \emph{Neural Networks}, 2021.

\bibitem[Xu and Mandic(2020)]{xu2020recurrent}
Y.~L. Xu and D.~P. Mandic.
\newblock Recurrent graph tensor networks.
\newblock \emph{arXiv preprint arXiv:2009.08727}, September 2020.

\bibitem[Xu et~al.(2020)Xu, Konstantinidis, and Mandic]{xu2020multigraph}
Y.~L. Xu, K.~Konstantinidis, and D.~P. Mandic.
\newblock Multi-graph tensor networks.
\newblock In \emph{Proceedings of the First Workshop on Quantum Tensor Networks
  in Machine Learning, 34th Conference on Neural Information Processing Systems
  (NeurIPS 2020)}, 2020.

\bibitem[Zhang et~al.(2018)Zhang, Yang, Chen, and Li]{zhang2018survey}
Q.~Zhang, L.~T Yang, Z.~Chen, and P.~Li.
\newblock A survey on deep learning for big data.
\newblock \emph{Information Fusion}, 42:\penalty0 146--157, 2018.

\bibitem[Zhang et~al.(2017)Zhang, Guo, Dong, He, Xu, and
  Chen]{zhang2017cautionary}
S.~Zhang, B.~Guo, A.~Dong, J.~He, Z.~Xu, and S.~X. Chen.
\newblock Cautionary tales on air-quality improvement in {B}eijing.
\newblock \emph{Proceedings of the Royal Society A: Mathematical, Physical and
  Engineering Sciences}, 473\penalty0 (2205):\penalty0 20170457, 2017.

\end{thebibliography}

\end{document}